%% file: main.tex
\documentclass[runningheads]{llncs}

\usepackage{eccv}

\usepackage{eccvabbrv}
\input{preamble}
\usepackage{graphicx}
\usepackage{booktabs}
\usepackage{multirow}
\usepackage[accsupp]{axessibility}  

\usepackage{hyperref}
\usepackage[final]{pdfpages}
\usepackage{orcidlink}
\usepackage[nottoc]{tocbibind}
\usepackage{marginnote}
\usepackage{textcomp}

\begin{document}

\title{HiFi-123: Towards High-fidelity One Image to 3D Content Generation}

\titlerunning{HiFi-123}

\author{
Wangbo Yu\inst{1,2}\orcidlink{0000-0003-4387-8967} \and
Li Yuan\inst{1,2}\orcidlink{0000-0002-2120-5588} \and
Yan-Pei Cao\thanks{Corresponding Authors.}$^{\dagger}$ \inst{3}\orcidlink{0000-0002-0416-4374} \and
Xiangjun Gao\inst{4}\orcidlink{0009-0003-6177-4413} \and \\
Xiaoyu Li\inst{3}\orcidlink{0000-0003-2588-1687} \and 
Wenbo Hu\inst{3}\orcidlink{0000-0001-6082-4966} \and
Long Quan\inst{4}\orcidlink{0000-0002-0329-9437} \and \\
Ying Shan\inst{3}\orcidlink{0000-0001-7673-8325} \and
Yonghong Tian$^{\star}$\inst{1,2}\orcidlink{0000-0002-2978-5935}
}

\authorrunning{Yu et al.}

\institute{
School of Electronic and Computer Engineering, Peking University \and
Peng Cheng Laboratory \and
Tencent AI Lab \and
Hong Kong University of Science and Technology
}

\maketitle
\input{sec/0_abstract}  
\footnotetext[4]{Now at VAST.}
\input{sec/1_intro}

\input{sec/2_relatedworks}
\input{sec/3_method}

\input{sec/4_experiment}
\input{sec/5_conclusion}

%
%

\input{main.bbl}
\bibliographystyle{splncs04}

\clearpage 
\begin{center}
\section*{\centering Supplementary Material}
\end{center}
\setcounter{section}{0} 
\setcounter{figure}{0} 
\setcounter{table}{0} 

\input{sec_supp/pre}
\addcontentsline{toc}{chapter}{Broad impact}
\input{sec_supp/impact}
\addcontentsline{toc}{chapter}{More implementation details of image-to-3D generation}
\input{sec_supp/implementation}
\addcontentsline{toc}{chapter}{Additional ablation studies}
\input{sec_supp/ablation}

\addcontentsline{toc}{chapter}{More results}
\input{sec_supp/moreresults}

\clearpage  
\end{document}

%% file: preamble.tex
\usepackage[dvipsnames]{xcolor}

\usepackage{bm}
\usepackage{graphicx}
\usepackage{amsfonts}       
\usepackage{amsmath,amssymb}
\usepackage{graphicx}
\usepackage{mathtools}
\usepackage{algorithm}
\usepackage{algorithmic}
\usepackage{amsthm}
\usepackage{mathrsfs} 
\usepackage{amsmath} 
\usepackage{booktabs} 
\input{math_commands.tex}

\usepackage{cuted}

%% file: math_commands.tex
\usepackage{amsmath,amsfonts,bm}

\def\eqref#1{equation~\ref{#1}}

\def\1{\bm{1}}

\def\mI{{\bm{I}}}

\DeclareMathAlphabet{\mathsfit}{\encodingdefault}{\sfdefault}{m}{sl}
\SetMathAlphabet{\mathsfit}{bold}{\encodingdefault}{\sfdefault}{bx}{n}

\def\gL{{\mathcal{L}}}

\def\gU{{\mathcal{U}}}



%% file: sec/0_abstract.tex
\begin{abstract}
\vspace{-8pt}
Recent advances in diffusion models have enabled 3D generation from a single image. However, current methods often produce suboptimal results for novel views, with blurred textures and deviations from the reference image, limiting their practical applications. In this paper, we introduce \textbf{HiFi-123}, a method designed for high-fidelity and multi-view consistent 3D generation. Our contributions are twofold: First, we propose a Reference-Guided Novel View Enhancement (RGNV) technique that significantly improves the fidelity of diffusion-based zero-shot novel view synthesis methods. Second, capitalizing on the RGNV, we present a novel Reference-Guided State Distillation (RGSD) loss. When incorporated into the optimization-based image-to-3D pipeline, our method significantly improves 3D generation quality, achieving state-of-the-art performance. Comprehensive evaluations demonstrate the effectiveness of our approach over existing methods, both qualitatively and quantitatively. 
Video results are available on the \href{https://drexubery.github.io/HiFi-123/}{project page}.
\end{abstract}

%% file: sec/1_intro.tex
\section{Introduction}
\label{sec:introduction}
The generation of 3D digital content is a fundamental task in computer vision and computer graphics with applications in robotics, virtual reality, and augmented reality. Producing such 3D content often demands proficiency in specialized software tools, setting a high threshold in terms of skill and cost. An alternative approach is through 3D digitization, which often relies on a large set of multi-view images and their corresponding camera poses; however, acquiring such data is challenging. A more ambitious approach is to construct 3D content from only a single image, whether obtained from the web or generated. While humans can intuitively infer 3D shapes and textures from 2D images, creating 3D assets from a single image using computer vision techniques is difficult due to the limited 3D cues and ambiguities of a single viewpoint.

\input{images_tex/teaser}
Recent advances in diffusion models, trained on web-scale 2D image datasets, have led to significant improvements in text-to-image (T2I) generation~\cite{ramesh2021zero,rombach2022high,balaji2022ediffi,nichol2021glide}. 
By leveraging the 3D priors inherent in T2I models, methods such as~\cite{poole2022dreamfusion,wang2023prolificdreamer} have utilized score distillation sampling (SDS) to achieve notable results in text-to-3D generation. This progress has also influenced the image-to-3D domain, with works such as~\cite{melas2023realfusion, xu2023neurallift, tang2023make} employing SDS loss combined with reference-view pixel losses to optimize neural representations from a single image. While these optimization-based image-to-3D methods can produce reasonable 3D structures from a single viewpoint, the visual quality in novel views often lacks fidelity, exhibiting inconsistencies with the reference image and oversmoothed textures, as shown in Fig.~\ref{fig:comp_3drec}.  The primary challenge arises when, for novel views outside the reference view, the optimization becomes overly reliant on, and thus closely tied to, the inferred text prompt of reference image. These text prompts, even those derived through textual inversion~\cite{gal2022image,burgess2023viewpoint}, often fail to capture the full visual details of the reference image, leading to inconsistent optimization results in novel views. What's more, the strong CFG guidance present in the SDS loss~\cite{poole2022dreamfusion} further amplifies the issue. These problems not only compromise the realism of the generated content but also limit its potential for broader applications.

Apart from optimization-based image-to-3D generation, Zero-1-to-3~\cite{liu2023zero} introduced an approach that demonstrated the efficacy of fine-tuning T2I models for zero-shot novel view synthesis, highlighting their ability to produce novel views in an optimization-free manner.
However, models like Zero-1-to-3 require fine-tuning on synthetic multi-view datasets~\cite{deitke2023objaverse}, which can lead to a noticeable degradation in model performance, particularly in generating unnatural and low-quality novel views that deviates from the reference image, as shown in Fig.~\ref{fig:comp_zero123}. 

In this work, we aim to enhance the fidelity and consistency for both zero-shot novel-view synthesis and optimization-based image-to-3D generation, endowing generation of photo-realistic 3D assets. To this purpose, we devise a method that can simultaneously generates consistent novel views from a single image while maintaining high image quality. Our primary insight lies in the application of the diffusion-based image inversion technique~\cite{song2020denoising} to retain the detailed structure and textures of a specific object, enabling the generation of novel views and the subsequent 3D representation of the object with consistent details. One of the key insight is we observe that by integrating depth information into the DDIM inversion~\cite{song2020denoising} and the sampling process based on a depth-conditioned stable diffusion model~\cite{SDD}, the reconstruction quality of the object is significantly improved and near perfect (illustrated in the Supplementary). Leveraging this observation, we introduce \textbf{HiFi-123}, a method that, while intuitive, effectively generates high-fidelity novel views and 3D content from a single reference image. 
Specifically, we design a Reference-Guided Novel View Enhancement (RGNV) pipeline in which both the reference image and a ``coarse'' estimation of the target novel view are inverted and reconstructed simultaneously, with the inversion process capturing fine details of the reference image, and the sampling process transferring texture details to the coarse novel view.
This RGNV pipeline can be seamlessly integrated into the recent zero-shot novel view synthesis methods~\cite{liu2023zero,liu2023syncdreamer}. Moreover, the inversion process's unique properties enable us to re-formulate and re-derive the SDS loss~\cite{poole2022dreamfusion}, resulting in a Reference-Guided State Distillation (RGSD) loss that is easy to implement and efficient to optimize. As a result, we can also achieve high-fidelity in optimization-based image-to-3D generation that significantly exceeds prior techniques.

We comprehensively evaluated \textbf{HiFi-123} on both zero-shot novel view synthesis and image-to-3D generation tasks. Both qualitative and quantitative results indicate that our approach excels in generating high-fidelity and consistent novel views from a single reference image and further produces high-quality 3D content. Compared to state-of-the-art approaches, our method shows significant improvements in visual quality, marking an important step towards more accessible and democratized 3D content creation.

In summary, the main contributions of our work are two-fold:
\begin{itemize}
\item
We introduce a Reference-Guided Novel View Enhancement (RGNV) pipeline grounded in a depth-based DDIM inversion. This approach can function as a plug-and-play module to improve the fidelity of results derived from diffusion-based zero-shot novel view synthesis methods.
\item
Leveraging the RGNV pipeline, we present a novel Reference-Guided State Distillation (RGSD) loss. When incorporated into the optimization-based image-to-3D framework, it significantly enhances the quality of 3D generation, achieving state-of-the-art performance.
\end{itemize}

%% file: images_tex/teaser.tex
\begin{figure*}
\centering
    \includegraphics[width=\linewidth]{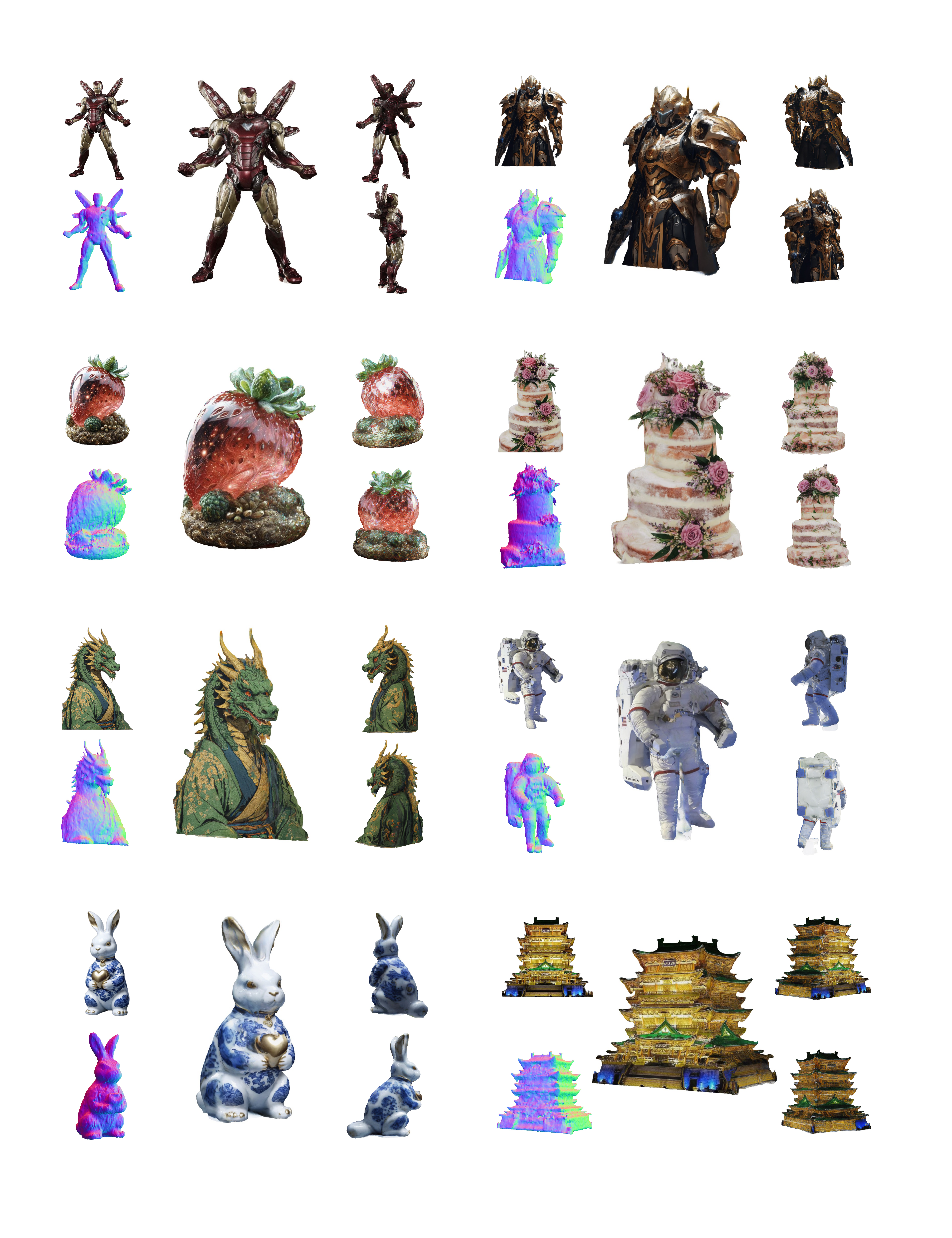}
    \caption{\textbf{HiFi-123} is capable of generating high-fidelity 3D content from a single reference image.  In each block above, we display the reference image (top left corner) along with the rendered novel views and normal of the generated 3D content. The presented novel views demonstrate that our approach maintains consistency and high-fidelity  with the reference image, even in views significantly deviating from the reference view.}
\end{figure*}

%% file: sec/2_relatedworks.tex
\section{Related Work}
\label{related}
\subsection{Optimization-based image-to-3D generation.}
Based on the powerful text-to-image diffusion models~\cite{nichol2021glide,ramesh2022hierarchical,saharia2022photorealistic,rombach2022high} in recent years, text-to-3D
generation has also made great progress.
DreamFields~\cite{jain2022zero} uses aligned image and text models to optimize NeRF~\cite{mildenhall2021nerf} without 3D shape or multi-view data. 
DreamFusion~\cite{poole2022dreamfusion} proposes a Score Distillation Sampling (SDS) method that replaces CLIP loss from DreamField with a loss derived from the distillation of a 2D diffusion model to optimize a parametric NeRF model, which becomes a paradigm for 3D generation using 2D diffusion. 
To improve the text-to-3D generation results, Magic3D~\cite{lin2023magic3d} builds upon DreamFusion that introduces several design choices like coarse-to-fine optimization, using Instant NGP representation in the coarse stage and 3D mesh representation in the fine stage. Fantasia3D~\cite{chen2023fantasia3d} further disentangles the modeling of geometry and appearance, and ProlificDreamer~\cite{wang2023prolificdreamer} proposes to modify score distillation sampling to variational score distillation which models the 3D parameters as a random variable instead of a constant. 
Apart from text-to-3D generation, 3D generation based on a single image using diffusion models (image-to-3D) has also made rapid progress. NeuralLift-360~\cite{xu2023neurallift} learns to recover a 3D object from a single reference image with CLIP-guided diffusion prior. In addition to using the SDS loss for distillation, RealFusion~\cite{melas2023realfusion} and NeRDi~\cite{deng2023nerdi} also adopt textual inversion to condition the diffusion model on a prompt with a token inverted by the reference image. Recently, Make-It-3D~\cite{tang2023make} employs textured point clouds as the representation in the fine stage to achieve high-quality results, Magic123~\cite{qian2023magic123} and DreamCraft3D~\cite{sun2023dreamcraft3d} suggests using an additional 3D diffusion prior trained on large-scale multi-view dataset for score distillation sampling. These methods often suffer from inconsistency between reference view and novel views.
\subsection{Diffusion-based zero-shot novel view synthesis}
Trained on large-scale 2D image datasets, the 2D text-to-image diffusion models could generalize to unseen scenes and different viewing angles that could be used for distilling 3D assets. However, due to the data bias of 2D images, e.g., most images are captured from front views, the 2D diffusion model may lack multi-view knowledge for 3D generation. Some efforts have been made to train the diffusion with 3D awareness. 3DiM~\cite{watson2022novel} and Zero-1-to-3~\cite{liu2023zero} present viewpoint-conditioned diffusion model for novel view synthesis trained on multi-view images. Utilizing large-scale 3D data, Zero-1-to-3 achieves zero-shot generalization ability to unseen images. One-2-3-45~\cite{liu2023one} uses the model from Zero-1-to-3 to generate multi-view images from the input view and leverage the generated results for 3D reconstruction. The recently works~\cite{burgess2023viewpoint, tang2023mvdiffusion, liu2023syncdreamer} try to generate multiview consistent images from a single view~\cite{liu2023syncdreamer}. However, these methods usually produce lower-quality results compared with the input view, limiting their broader applications.

%% file: sec/3_method.tex
\vspace{-.3cm}
\section{Methodology}
\vspace{-.1cm}
\subsection{Preliminary}
\textbf{Diffusion models.} 
A diffusion model consists of a forward process $q$ and a reverse process $p$. In the forward process, starting from a clean data $\bm{x}_0 \sim q_0(\bm{x}_0)$, noise is gradually added to the data point $\bm{x}_0$ to construct noisy state at different time steps, formulated as $\bm{x}_t = \alpha_t\bm{x}_0+\sigma_t\epsilon$, 
where $\alpha_t$ and $\sigma_t$ are hyper-parameters satisfying $\alpha^2_t + \sigma^2_t = 1$, $\epsilon\sim\mathcal{N}(\bm{0},\mI)$. 
The reverse process $p_{\phi}$ is defined by removing noise added on the clean data using a U-Net noise predictor $\epsilon_\phi$. In text-to-image diffusion models~\cite{saharia2022photorealistic,ramesh2022hierarchical,rombach2022high}, $\epsilon_\phi$ is trained by minimizing the score matching objective:
\begin{equation}
\label{eq:diffusion_loss}
\gL_{\text{Diff}}(\phi) = \mathbb{E}_{t\sim\gU(0,1}[w(t)\|\epsilon_\phi(\bm{x}_t;y,t) - \epsilon\|_2^2],
\end{equation}
where $w(t)$ is a time-dependent weighting function and $y$ is conditional text embedding. 
To balance the quality and diversity of the generated images, classifier-free guidance (CFG~\cite{ho2022classifier}) is adopted to modify the estimated noise as a combination of conditional and unconditional output: $\hat\epsilon_\phi(\bm{x}_t;y, t)= (1+s)\epsilon_\phi(\bm{x}_t;y, t) - s \epsilon_\phi(\bm{x}_t; t)$, where $s>0$ is the guidance scale. Increasing the guidance scale typically enhances the alignment between text and image, but at the cost of reduced diversity.
\input{images_tex/pipeline}

\noindent\textbf{DDIM inversion.} 
In the reverse process $p_{\phi}$, diffusion models often utilize deterministic DDIM sampling~\cite{song2020denoising} to speed up inference. DDIM sampling converts random noise $\bm{x}_T$ into clean data $\bm{x}_0$ over a sequence of discrete time steps, from $t=T$ to $t=1$, formulated as: $\bm{x}_{t-1} = (\alpha_{t-1}/\alpha_t)(\bm{x}_t-\sigma_t\epsilon_\phi)+\sigma_{t-1}\epsilon_\phi$. 
In contrast, DDIM inversion~\cite{dhariwal2021diffusion,song2020denoising} is a forward process that gradually converts a clean data $\bm{x}_0$ back to a noisy state $\bm{x}_T$ using denoising U-Net $\epsilon_\phi$. From $t=1$ to $t=T$, we have $\bm{x}_{t} = (\alpha_{t}/\alpha_{t-1})(\bm{x}_{t-1}-\sigma_{t-1}\epsilon_\phi)+\sigma_{t}\epsilon_\phi$. In the case of unconditional generation, the DDIM inversion process $q_\phi$  is completely consistent with the sampling process $p_\phi$, so that the original data $\bm{x}_0$ can be precisely reconstructed by applying DDIM sampling on the inverted $\bm{x}_T$. However, for the text-conditioned generation with classifier-free guidance, the two processes are not consistent and the reconstruction quality will significantly decrease~\cite{mokady2023null}.

\noindent\textbf{Score distillation sampling (SDS).}
SDS~\cite{poole2022dreamfusion} is an optimization method commonly used in recent text-to-3D generation~\cite{poole2022dreamfusion,wang2023score,lin2023magic3d,metzer2023latent,wang2023score,chen2023fantasia3d} and image-to-3D generation methods~\cite{melas2023realfusion,qian2023magic123,xu2023neurallift,tang2023make}. The core idea of SDS is to distill prior knowledge from pre-trained T2I models by minimizing:
\begin{equation}
\begin{split}
    \gL_{\text{SDS}}(\theta)
    = \mathbb{E}_t &\left[(\sigma_t / \alpha_t) w(t) \text{KL}(q(\bm{x}_t|g(\theta,c); y,t) \| p_{\phi}(\bm{x}_t;y,t ))\right],
    \label{eq:sds}
\end{split}
\end{equation}
where $\theta$ denotes the parameters of a trainable 3D representation (e.g., NeRF~\cite{mildenhall2021nerf} or DMTet~\cite{shen2021dmtet}) and $g(\theta,c)$ is a rendered image given a camera pose $c$. By minimizing the KL divergence between distributions of noisy renderings and denoised images at different time steps, the 3D representation will be optimized to match the distribution of the images synthesized by the text-to-image diffusion model. In practice, the gradients of Eq.~\ref{eq:sds} is approximated by~\cite{poole2022dreamfusion}:
\begin{equation}
\nabla_{\theta} \gL_{\text{SDS}}(\theta) \approx \mathbb{E}_{t, \epsilon}\left[w(t)\left(\epsilon_\phi(\bm{x}_t; y, t)  - \epsilon\right) {\partial \bm{x}_0 \over \partial \theta}\right].
\label{eq:sdsgrad}
\end{equation}
Although optimizing with SDS loss can result in overall reasonable geometry, the generated 3D model often exhibits over-saturated colors and over-smoothed textures~\cite{poole2022dreamfusion}, which could lead to inconsistent results compared with the reference image when applied to image-to-3D generation tasks.

\subsection{Reference-Guided Novel View Enhancement for zero-shot novel view synthesis}
\label{sec:3.2}
Given a reference image, previous diffusion-based zero-shot novel view synthesis methods~\cite{liu2023zero,poole2022dreamfusion} prone to produce degraded and inconsistent results in novel views compared with the reference view. 
To tackle this problem, we propose a Reference-Guided Novel view Enhancement (RGNV) pipeline to transfer the detailed textures of the reference image to the coarse novel view.
Our pipeline is built upon a discovery that incorporates depth map into the DDIM inversion and sampling process using a depth-conditioned diffusion model~\cite{SDD} will near perfectly reconstruct the reference image, achieving comparable performance with optimization-based inversion~\cite{mokady2023null} (discussed in the Supplementary). With this discovery, we can obtain the initial noise and the reverse processes that can faithfully reconstruct the detailed textures of the reference image in a zero-shot manner. 
Then, inspired by the progressive generation property of the reverse process where the geometry structure emerges first at the early denoising steps while texture details appear at the late denoising steps, 
we design a dual-branch pipeline to transfer fine textures of the reference image to the coarse novel views.

 As shown in Fig.~\ref{pipeline}, our pipeline performs DDIM inversion and sampling on both the reference image and coarse novel view simultaneously. In the forward process $q_\phi$, we separately map the reference image and coarse novel view back to the initial noisy state, denoted as $\bm{x}^r_T$ and $\bm{x}_T$, with $t=T$ steps' DDIM inversion. Subsequently, in the reverse process $\Tilde{p}_\phi$ (differs from the the regular reverse process $p_\phi$) of the pipeline, we first perform DDIM sampling separately on the two states to denoise them for $t=T-l$ steps, where coarse geometry structure has emerged. Then, in the following $t=l$ denoising steps where fine textures will gradually appear in the reference image branch, inspired by recent works on consistent video generation~\cite{wu2022tune,cao_2023_masactrl}, we replace the $K,V$ matrices of denoising U-Net's self-attention in the coarse novel view branch with the corresponding matrices $K^r,V^r$ in the reference image branch, which we term as attention injection. Through attention injection, fine textures of the reference image will be transferred to the coarse novel view. Thanks to the nearly perfect reconstruction quality of depth-based DDIM inversion, the inversion process and the sampling process are nearly consistent at every time step, we can thus simplify the pipeline to directly invert the two inputs for $t=l$ steps, and then symmetrically adopt $t=l$ denoising steps with attention injection to propagate textures of the reference view to the coarse novel view. 

The RGNV pipeline can serve as a plug-and-play method for enhancing the quality of diffusion-based zero-shot novel view synthesis methods~\cite{liu2023zero,liu2023syncdreamer}, as shown in Fig.~\ref{fig:comp_zero123}. We also demonstrate it can improve optimization-based image-to-3D generation in the following section.

\input{images_tex/pipeline3d}
\subsection{Reference-Guided State Distillation for image-to-3D generation.}
As shown in Fig.~\ref{pipeline3d}, we adopt a coarse-to-fine optimization strategy to create 3D content from a single reference image. In the coarse stage, we use hybrid SDS loss provided by a 2D image diffusion model~\cite{DF} and a 3D novel view synthesis diffusion model~\cite{liu2023zero} to optimize a coarse Instant NGP~\cite{muller2022instant}. The reference view reconstruction loss, depth loss, and normal loss are also involved to supervise training. 
As shown in Fig.~\ref{fig:rgsd}.~(a), after the coarse stage training, the resulting 3D representation already possesses reasonable geometry and colors. However, it suffers from over-smoothing and over-saturation of textures produced by the SDS loss. 

In the refine stage, we convert the implicit NeRF into an explicit DMTet representation~\cite{shen2021deep} with learnable parameter $\theta$ for higher rendering resolution and efficient training. In particular, we fix the geometry of the DMTet and focus on refining texture details in this stage. 
Several works~\cite{tang2023make,qian2023magic123} continue optimizing the texture details using SDS loss in the refine stage. 
Nonetheless, it can be observed in Eq.~\ref{eq:sds} that SDS loss leads to an optimization direction that forces the forward process $q$ of rendered novel views to approach the distribution of the reverse process $p_\phi$ of text-to-image generation. 
Due to the ambiguity of the inferred text descriptions and the large CFG guidance, the optimized novel views are often inconsistent with the reference image, as shown in Fig.~\ref{fig:rgsd}.~(b). 

To address the inferior textures caused by SDS loss and ensure high fidelity in novel views, we integrate our proposed RGNV pipeline into the refine stage.
One naive approach would be to randomly render coarse novel views, utilize the RGNV pipeline for enhancement, and subsequently apply reconstruction loss using the enhanced images. We refer to it as image loss. As shown in Fig.~\ref{fig:rgsd}.~(c), we found this approach produces oversmoothed textures, as even slight inconsistencies in the overlapping areas between enhanced images can accumulate and lead to blurry optimization results. 
Inspired by SDS loss~(Eq.~\ref{eq:sds}) that distills from the noisy states of text-to-image generation process for 3D generation, 
we propose a Reference-Guided State Distillation (RGSD) loss to distill from the generation process
of our RGNV pipeline for high-fidelity and consistent texture synthesis. 
Specifically, we construct a series of optimization targets using intermediate states from the RGNV pipeline,
the resulting objective can be formulated as:
\begin{equation}
    \min_{\theta}\!~\mathbb{E}_{t\sim\gU[0,l/T)}[\text{KL}(q_{\phi}(\bm{x}_t|g(\theta,c);\! y,\!m,\!t) \| \Tilde{p}_{\phi}(\bm{x}_t;\!y,\!m,\!r,\!t ))],
    \label{eq:enhacne}
\end{equation}
where $m$ denotes the conditioned depth map, $r$ denotes the reference image, $q_{\phi}$ and $\Tilde{p}_{\phi}$ are the inversion and sampling process of the RGNV pipeline. 
Compared with the SDS loss (Eq.~\ref{eq:sds}) that relies on the inferred high-level text prompts for optimization, 
this improved objective forces noisy states $\bm{x}_t$ of the coarse novel views to approach their enhanced states $\Tilde{\bm{x}}_t$ produced by the reference-conditioned RGNV pipeline, 
ensuring that the supervision from the reference image can cover all the camera views, thereby endowing an accurate optimization direction towards the distribution of the 3D object that is consistent with the reference image. 

\input{images_tex/SDS}
Since $q_{\phi}$ and $\Tilde{p}_{\phi}$ are deterministic processes given a specific reference image, we can solve Eq.~\ref{eq:enhacne} using a distance metric~\cite{huang2020probability} such as L2 distance. In this way, Eq.~\ref{eq:enhacne} can be simplified as:
\vspace{-.1cm}
\begin{equation}
\label{eq:l2}
 \gL_{\text{RGSD}}(\theta)
    = \mathbb{E}_{t\sim\gU[0,l/T)}[\|\bm {x}_t - \Tilde{\bm{x}}_t\|_2^2].
\end{equation}
\vspace{-.001cm}
There are two inefficiencies in solving this objective. 
First, it requires multiple estimation of the U-Net $\epsilon_\phi$ to get the optimization target $\Tilde{\bm{x}}_t$. 
As depicted in Fig.~\ref{pipeline}, we need to invert the rendered novel view $\bm {x}_0 = g(\theta,c)$ into a noisy state $\bm{x}_l \sim q_\phi(\bm{x}_l)$ with $l$-step DDIM inversion, then perform with attention injection to denoise $\bm{x}_l$ to an enhanced latent $\Tilde{\bm{x}}_t \sim \Tilde{p}_\phi(\Tilde{\bm{x}}_t)$ and detach it from the computation graph to make it the final optimization target.
To accelerate training, we pre-select two fixed camera views and derive their $\Tilde{\bm{x}}_0$ states through the RGNV pipeline, using them as the optimization target at the $t=0$ time step. During training, we alternate between sampling $t=0$ to optimize the pre-defined $\Tilde{\bm{x}}_0$ states with fixed camera poses and sampling $t\sim\gU(0,l/T)$ to optimize the intermediate $\Tilde{\bm{x}}_t$ states with random camera poses. We found this leads to faster convergence compared with SDS loss, and results in superior results in novel views.
Second, as shown in Eq.~\ref{eq:enhacne}, unlike regular forward process $q$ where the gradients of $\bm{x}_t=\alpha_t\bm{x}_0+\sigma_t\epsilon$ can be efficiently calculated, it requires multiple forward-pass of U-Net $\epsilon_\phi$ to get $\bm{x}_t$ in DDIM inversion $q_\phi$, in which the gradient is expensive to compute. We therefore turn to an approximate solution to compute gradients of $\bm{x}_t$. Instead of constructing $\bm{x}_t$ by adding noise to $\bm{x}_0$ step by step using DDIM inversion, we use the deterministic noise $\Tilde{\epsilon}_t$ predicted from $\Tilde{\bm{x}}_t$ in the DDIM sampling process to construct noisy states for $\bm{x}_0$, so that the resulted $\bm{x}_t = \alpha_t\bm{x}_0 + \sigma_t\Tilde{\epsilon}_t$ will have the same noisy level with $\Tilde{\bm{x}}_t$. By this means, the gradients of $\bm{x}_t$ can be efficiently computed. 

As shown in Fig.~\ref{fig:rgsd}.~(d), optimizing with RGSD loss effectively resolves the issues of inconsistent color and oversmoothed textures, resulting in consistent appearance with the reference image.
Please refer to Supplementary for a summarized algorithm of RGSD loss and more implementation details of the two training stages.

%% file: images_tex/pipeline.tex
\begin{figure}[t]
  \centering
  \includegraphics[width=0.95\textwidth]{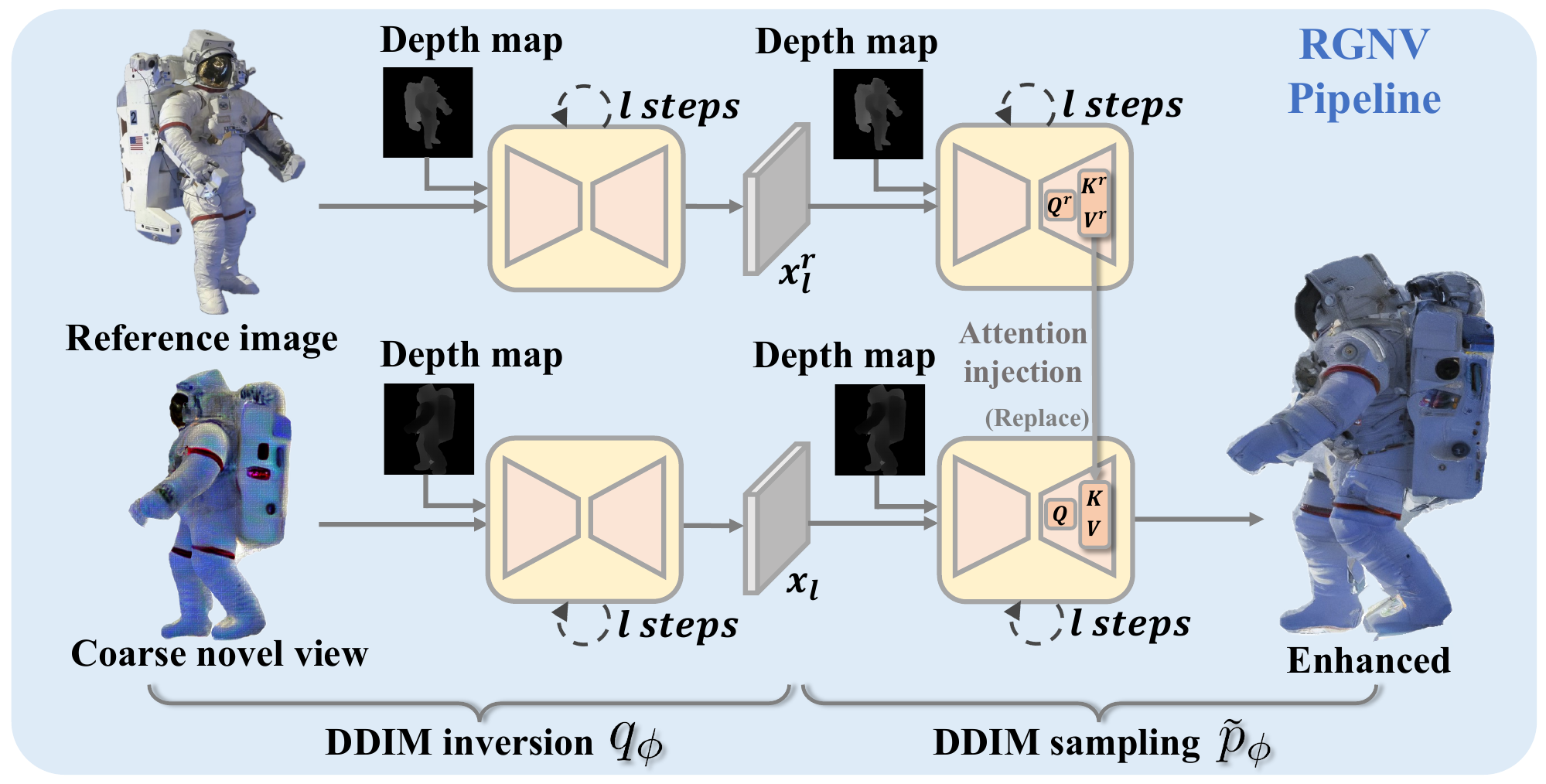}
  \vspace{-.1cm}
  \caption{Illustration of the RGNV pipeline. It performs depth-based DDIM inversion and sampling on both the reference image and coarse novel view, and utilizes attention injection to transfer detail textures from the reference image to the coarse novel view.}
  \vspace{-.6cm}
\label{pipeline}
\end{figure}

%% file: images_tex/pipeline3d.tex
\begin{figure*}[t]
  \centering
  \includegraphics[width=1.\textwidth]{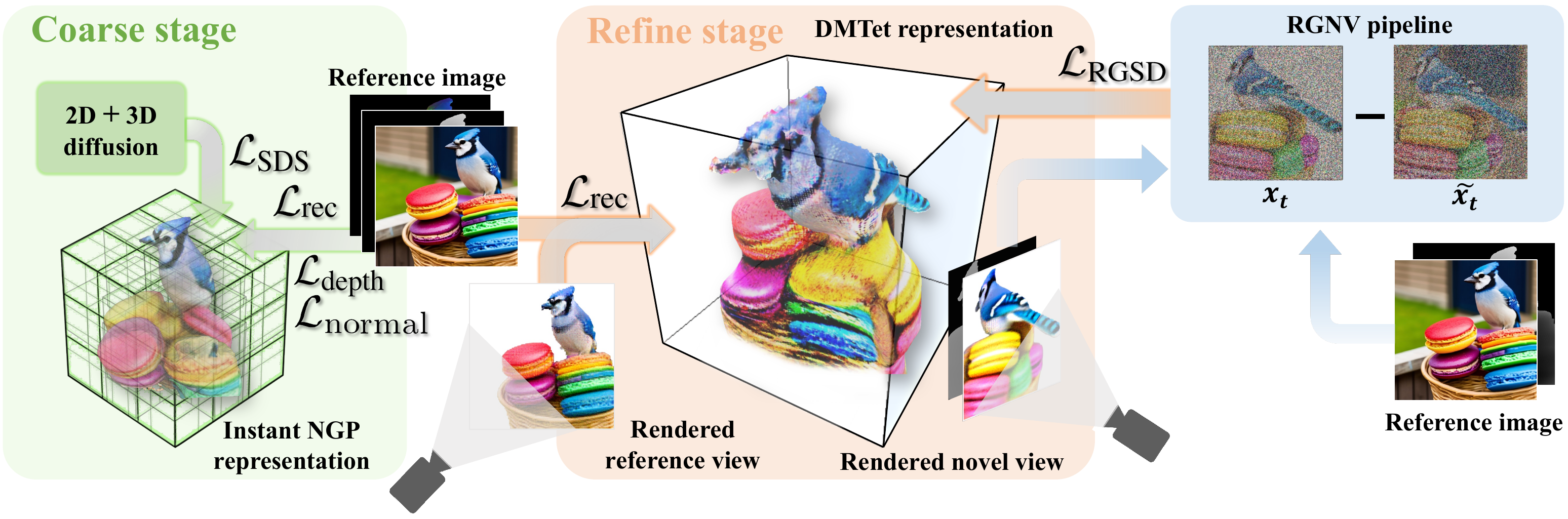}
  \vspace{-.6cm}
  \caption{Image-to-3D generation pipeline. We utilize two stages to generate high-fidelity 3D contents. In the coarse stage, we optimize an Instant-NGP representation using SDS loss, reference view reconstruction loss, depth loss, and normal loss. In the refine stage, we export DMTet representation and use our proposed RGSD loss to supervise training.}
  \vspace{-.6cm}
\label{pipeline3d}
\end{figure*}

%% file: images_tex/SDS.tex
\begin{figure}[t]
  \centering
  \includegraphics[width=.98\textwidth]{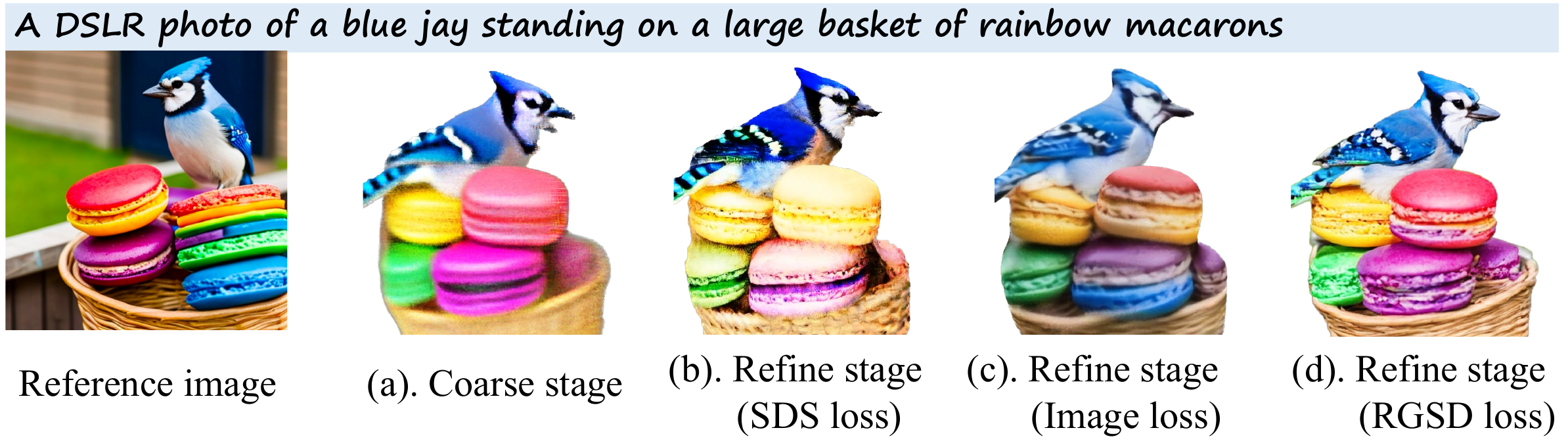}
  \vspace{-0.2cm}
  \caption{Comparison of using different losses in the refine stage.}
  \vspace{-.5cm}
\label{fig:rgsd}
\end{figure}

%% file: sec/4_experiment.tex
\section{Experiments}
\subsection{Implementation details}
\noindent\textbf{Zero-shot novel view synthesis.}
For the RGNV pipeline, we use MiDaS~\cite{Ranftl2021} to estimate depth maps for both the reference image and coarse novel view, and normalize the depth map into $[-1,1]$ to align with the depth-conditioned SD model~\cite{SDD}. We adopt $T=50$ steps' DDIM inversion, and set $l=30$ for attention injection. 

\input{images_tex/comp_zero123}
\noindent\textbf{Image-to-3D generation.}
In the coarse stage, we use an Instant NGP~\cite{muller2022instant} representation optimized from 64 to 128 resolution. In the refine stage, we export DMTet~\cite{shen2021dmtet} and use a rendering resolution of 1024. For the RGSD loss implementation, we use $T=20$ steps' DDIM inversion with the attention injection start step set to $l=12$. The coarse stage training takes about 30 minutes, and the refine stage training takes 10 minutes, both tested on a 40G A100 GPU.


\subsection{Zero-shot novel view synthesis comparison}
\textbf{Baselines.} We use Zero-1-to-3~\cite{liu2023zero} and SyncDreamer~\cite{liu2023syncdreamer} as the baseline methods to assess our RGNV pipeline, both of which are  diffusion-based zero-shot novel view synthesis methods. 
Specifically, Zero-1-to-3 allows for explicit control over the generation of novel views through relative camera poses. SyncDreamer is capable of simultaneously generating 16 novel views from a single image, with pre-defined camera poses.

\noindent\textbf{Comparison on single view dataset.}  
We compare our method with the baselines using 400 images, including challenging real-world images and realistic images generated by a T2I model~\cite{DF}.
Fig.~\ref{fig:comp_zero123} presents the qualitative comparison with Zero-1-to-3, please refer to the Supplementary for qualitative comparison with SyncDreamer.
We found that although the novel views generated by the baselines exhibit reasonable geometry, their textures lack details and appear to be unreasonable, resulting in poor consistency with the reference image. In comparison, by applying the RGNV pipeline on the baselines, the fidelity and texture quality of the generated novel views are significantly improved. For quantitative evaluation, referring~\cite{tang2023make,qian2023magic123,xu2023neurallift}, we adopt contextual distance~\cite{mechrez2018contextual} and CLIP-similarity~\cite{radford2021learning} to measure the consistency between reference image and novel views. Since the baselines cannot generate images with background, to ensure a fair comparison, we mask out the background generated by our method when computing the metrics.
The results are listed in Tab.~\ref{tab:rgnv}, which reflects the effectiveness of the RGNV pipeline.
\input{tables/RGNV}

\noindent\textbf{Comparison on 3D dataset.}  
\label{sec:4.1}
For 3D evaluation, our evaluation dataset is the same with that of SyncDreamer~\cite{liu2023syncdreamer}, comprising of 30 objects from the Google Scanned Object dataset~\cite{downs2022google}, each with 16 rendered novel views for evaluation.
We adopt PSNR, SSIM~\cite{wang2004image} and LPIPS~\cite{zhang2018unreasonable} to quantitatively evaluate the novel view synthesis quality, the results are shown in Tab.~\ref{tab:rgnv}, validating that the RGNV pipeline helps to improve novel views synthesis quality. Qualitative results are displayed in the Supplementary.
\input{images_tex/comp_3drec}
\subsection{Image-to-3D generation comparison}
\textbf{Baselines.}
We compare our image-to-3D generation framework against three baselines: RealFusion~\cite{melas2023realfusion}, Make-It-3D~\cite{tang2023make} and Magic123~\cite{qian2023magic123}. RealFusion is a one-stage method that reconstructs NeRF representation from the reference image using L2 reconstruction loss and 2D SDS loss. Make-It-3D is a two-stage method that leverages point cloud representation in the second stage for training at higher resolution. 2D SDS loss is adopted in its two stages for geometry sculpturing and texture refining. Magic123 is also a two-stage method that uses both 2D SDS loss and 3D SDS loss provided by Zero-1-to-3~\cite{liu2023zero} to balance between geometry and texture quality. 

\noindent\textbf{Comparison on single view dataset.}
We firstly conduct comparisons against baseline methods on the aforementioned single view dataset.
Fig.~\ref{fig:comp_3drec} displays the qualitative comparison between our method and the baselines, where we showcase two novel views for each generated object. We also present a comparison of the normal map optimized by Magic123~\cite{qian2023magic123} and our method. It can be observed that, under the viewpoint that deviates significantly from the reference image, all the baseline methods fail to generate reasonable textures. The inconsistency is particularly evident at the boundaries between invisible and occluded areas, resulting in noticeable seams. In contrast, our method can maintain the same texture details as the reference image, which greatly improves the fidelity of the generated 3D assets. 
Please refer to the supplementary videos for a more comprehensive comparison.
For quantitative evaluation, 
except for adopting CLIP-similarity and contextual distance for evaluating novel views, we also use LPIPS~\cite{zhang2018unreasonable} to evaluate the reference view reconstruction quality. The results are reported in Tab.~\ref{tab:RGSD}, where the CLIP-similarity and contextual distance validates that our method can generate 3D objects with better 3D consistency. 

\input{tables/RGSD}
\input{images_tex/ablate_nvs}
\noindent\textbf{Comparison on 3D dataset.}  
Following the 3D evaluation settings in Sec.~\ref{sec:4.1}, we adopt the Google Scanned Object dataset~\cite{downs2022google} and use 30 objects for evaluation, and use PSNR, SSIM~\cite{wang2004image} and LPIPS~\cite{zhang2018unreasonable} to quantitatively evaluate the novel view synthesis quality. Referring~\cite{liu2023syncdreamer}, we also utilize the Chamfer Distance and Volume IoU to evaluate the generated geometry. Tab.~\ref{tab:RGSD} shows the quantitative results, which validates that our method is capable of generating 3D contents with better texture details as well as reasonable geometry. Qualitative results are in the Supplementary.
\subsection{Ablation study}
\noindent\textbf{Design space of the RGNV pipeline.}
There are two key designs in the RGNV  pipeline: the depth-based DDIM inversion and the attention injection. We qualitatively validate the effectiveness of these designs. 
As shown in Fig.~\ref{fig:ablate_nvs}, given a reference image and a generated coarse novel view, a naive approach to improve the novel view quality is  adopting SDEdit~\cite{meng2022sdedit}, which introduces random noise on the coarse novel view and denoise it to a clean image using pretrained diffusion model. However, we found enhancement results of SDEdit (Fig.~\ref{fig:ablate_nvs}.~(a)) presents color and textures inconsistent with the reference image, because it didn't make use of the reference information. In Fig.~\ref{fig:ablate_nvs}.~(b), performing RGNV without the depth condition~\cite{SDD} also leads to inconsistent enhanced results. The reason lies in that the regular DDIM  inversion (without depth condition) cannot precisely reconstruct the reference image (illustrated in the Supplementary), thus failing to transfer fine textures of reference image to the coarse novel view. Further, as shown in Fig.~\ref{fig:ablate_nvs}.~(c), directly using the depth-conditioned SD model~\cite{SDD} without reference attention injection also leads to view inconsistent results. In contrast, as shown in Fig.~\ref{fig:ablate_nvs}.~(d), with depth-based DDIM inversion that capture fine details of the reference image and attention injection that transfer fine textures to the coarse novel views, our RGNV pipeline can produce enhanced images consistent with the reference image. More ablations are presented in the supplementary.

\input{tables/ablation}
\input{images_tex/more}
\noindent\textbf{Effectiveness of the RGSD loss.}
We adopt a coarse-to-fine strategy for image-to-3D generation. In the refine stage, we propose a RGSD loss to improve texture quality and consistency. Qualitative results of the coarse stage and refine stage are shown in Fig.~\ref{fig:more}. It can be found that although the coarse stage can provide a reasonable geometry, its texture details are different from the reference image. Through refine stage optimization using our proposed RGSD loss, the texture of the novel views are significantly improved.
We further conduct quantitative ablation on datasets adopted in previous experiments, and evaluate the following settings: coarse stage, refine stage using SDS loss, and refine stage using RGSD loss, results are reported in Tab.~\ref{tab:ablation}. The results further demonstrate the effectiveness of refine stage training using RGSD loss, and validate that RGSD achieves better performance in enhancing texture quality and consistency than SDS loss.

%% file: images_tex/comp_zero123.tex
\begin{figure*}[t]
  \centering
  \includegraphics[width=1.\textwidth]{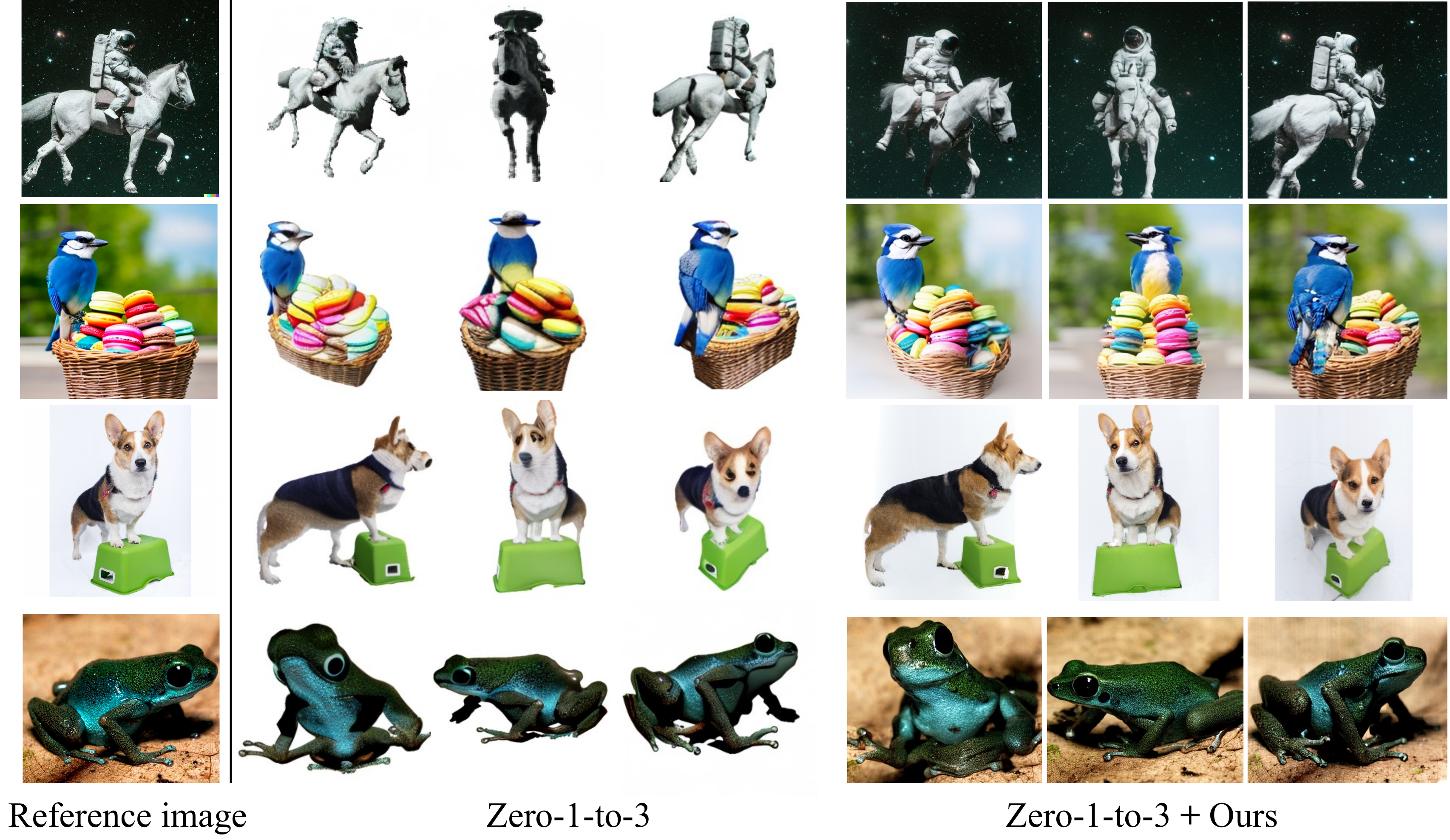}
  \vspace{-1em}
  \caption{Qualitative comparison with Zero-1-to-3~\cite{liu2023zero} on zero-shot novel view synthesis.``\textbf{+}Ours'' denotes enhanced by our RGNV pipeline. Our method helps to generate novel views with higher fidelity and finer texture details.}
  \vspace{-1em}
\label{fig:comp_zero123}
\end{figure*}

%% file: tables/RGNV.tex
\begin{table}[t]
  \caption{Comparison with Zero-1-to-3~\cite{liu2023zero} and SyncDreamer~\cite{liu2023syncdreamer} on single view dataset and 3D dataset. ``\textbf{+}Ours'' denotes enhanced by our RGNV pipeline.}
  \vspace{-.2cm}
  \label{tab:rgnv}
\centering
    \resizebox{.78\linewidth}{!}{
  \begin{tabular}{l|cc|ccc}
  \hline
& \multicolumn{2}{c|}{Single view dataset} & \multicolumn{3}{c}{3D dataset}\\
\hline
 Methods& Contextual$\downarrow$ & CLIP$\uparrow$ &PSNR$\uparrow$  & SSIM$\uparrow$ & LPIPS$\downarrow$ \\ \hline
Zero-1-to-3~\cite{liu2023zero}& 1.742 & 0.825& 18.95 & 0.782 & 0.163\\
Zero-1-to-3$+$Ours& $\mathbf{1.605}$ & $\mathbf{0.884}$& $\mathbf{20.45}$ & $\mathbf{0.810}$ & $\mathbf{0.149}$\\
SyncDreamer~\cite{liu2023syncdreamer}&1.709&0.851& 19.98 & 0.816 & 0.142\\
SyncDreamer$+$Ours&$\mathbf{1.598}$&$\mathbf{0.896}$& $\mathbf{21.08}$ & $\mathbf{0.849}$ & $\mathbf{0.123}$\\
\hline
\end{tabular}}
\vspace{-.4cm}
\end{table}

%% file: images_tex/comp_3drec.tex
\begin{figure*}[t]
  \centering
  \includegraphics[width=1.\textwidth]{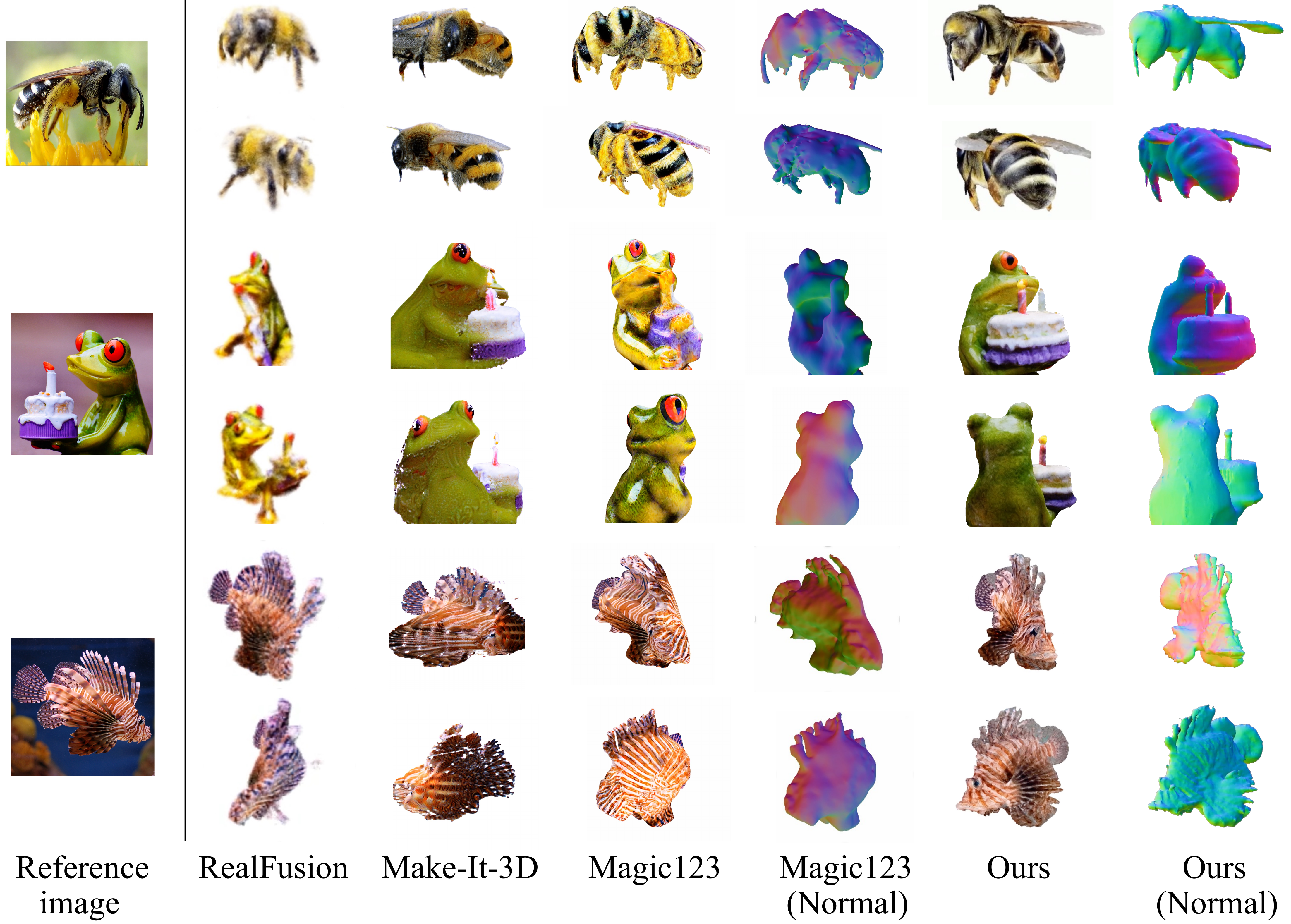}
  \vspace{-.4cm}
  \caption{Qualitative comparison with image-to-3D baselines. For each case, We show two novel views with a large angle from the reference image. It can be found that Our method outperforms baselines in maintaining texture details under significantly deviating viewpoints. Please refer to the video comparison in the Supplementary for more details.}
  \vspace{-.5cm}
\label{fig:comp_3drec}
\end{figure*}

%% file: tables/RGSD.tex
\begin{table}[t]
  \caption{Comparison with image-to-3D generation baselines on single view dataset and 3D dataset.}
  \vspace{-.2cm}
  \label{tab:RGSD}
\centering
    \resizebox{1.\linewidth}{!}{
  \begin{tabular}{l|ccc|ccccc}
  \hline
& \multicolumn{3}{c|}{Single view dataset} & \multicolumn{5}{c}{3D dataset}\\
\hline
 Methods& LPIPS$\downarrow$ &Contextual$\downarrow$ & CLIP$\uparrow$ & PSNR$\uparrow$ & SSIM$\uparrow$ & LPIPS$\downarrow$ & CD$\downarrow$ &IoU$\uparrow$ \\ \hline
RealFusion~\cite{melas2023realfusion}& 0.195 & 2.180& 0.767& 15.37 & 0.715 & 0.288&0.082&0.274\\
Make-It-3D~\cite{tang2023make} & 0.097 & 1.978& 0.898& 17.08 & 0.783 & 0.225&0.064&0.401\\
Magic123~\cite{qian2023magic123}& 0.085 & 1.882 & 0.883& 19.33 & 0.801 & 0.156&0.052&0.453\\
Ours& \textbf{0.081} & \textbf{1.627} & \textbf{0.916}& $\mathbf{23.68}$ & $\mathbf{0.875}$ & $\mathbf{0.101}$&\textbf{0.025}&\textbf{0.577}\\
\hline
\end{tabular}}
\end{table}

%% file: images_tex/ablate_nvs.tex
\begin{figure}[t]
  \centering
  \includegraphics[width=1.\textwidth]{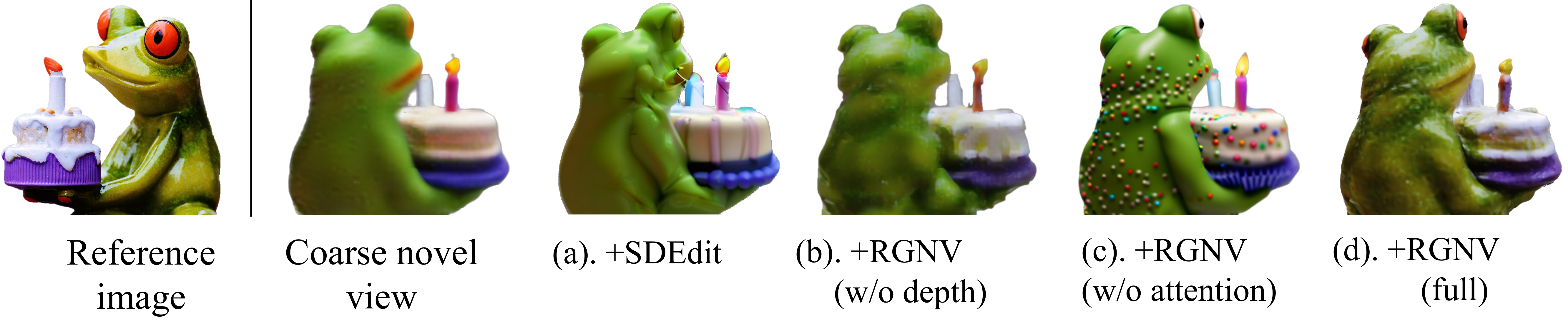}
  \vspace{-.5cm}
  \caption{Ablation on design space of the RGNV pipeline.}
  \vspace{-.5cm}
\label{fig:ablate_nvs}
\end{figure}

%% file: tables/ablation.tex
\begin{table}[t]
  \caption{Quantitative ablation on the effectiveness of RGSD loss.}
  \vspace{-.2cm}
  \label{tab:ablation}
\centering
    \resizebox{.85\linewidth}{!}{
  \begin{tabular}{l|cc|ccc}
  \hline
& \multicolumn{2}{c|}{Single view dataset} & \multicolumn{3}{c}{3D dataset}\\
\hline
 Settings&Contextual$\downarrow$ & CLIP$\uparrow$ & PSNR$\uparrow$ & SSIM$\uparrow$ & LPIPS$\downarrow$ \\ \hline
Coarse stage& 1.901& 0.836& 17.82 & 0.794 & 0.215\\
Refine stage (SDS loss)& 1.925 & 0.855& 19.25 & 0.810 & 0.188\\
Refine stage (RGSD loss)& \textbf{1.627} & \textbf{0.916}& $\mathbf{23.68}$ & $\mathbf{0.875}$ & $\mathbf{0.101}$\\
\hline
\end{tabular}}
\end{table}

%% file: images_tex/more.tex
\begin{figure}[t]
  \centering
  \includegraphics[width=.86\textwidth]{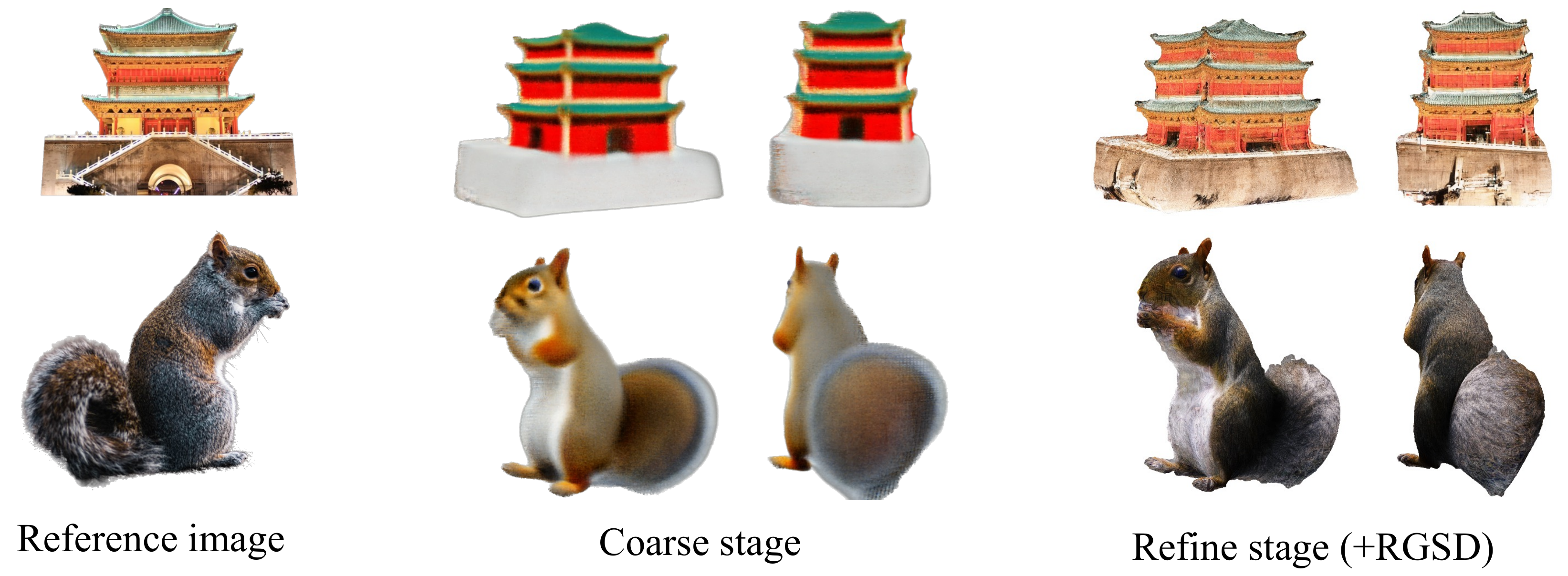}
  \vspace{-.2cm}
  \caption{Qualitative ablation on the effectiveness of RGSD loss.}
\vspace{-.5cm}
\label{fig:more}
\end{figure}

%% file: sec/5_conclusion.tex
\vspace{-.3cm}
\section{Conclusion and Discussion}
\vspace{-.2cm}
\textbf{Conclusion.} We introduce HiFi-123, a method that can be applied for generating high-fidelity novel views in a zero-shot manner as well as high-quality 3D contents. Our approach has two key contributions. Firstly, we propose an RGNV pipeline, which narrows the quality gap between synthesized and reference views in zero-shot novel view synthesis. Based on this pipeline, we further derive an RGSD loss to supervise and optimize 3D representations, resulting in highly realistic 3D assets. 

\noindent\textbf{Limitations.} 
 The RGNV pipeline currently requires a coarse novel view to provide an initial structure. This makes it more act as a plug-and-play module for existing zero-shot novel view synthesis methods~\cite{liu2023zero,liu2023syncdreamer}, and will inherit their generated wrong geometry. Pursuing a pure standalone approach for high-fidelity novel view synthesis remains a promising direction for future research. In addition, since a single reference image can provide very limited 3D cues, our image-to-3D framework may suffer from geometry ambiguity and fail to reconstruct reasonable invisible views.

%% file: sec_supp/pre.tex
In the supplementary material, we first discuss the broad impact of our method, then present more implementation details of image-to-3D generation, followed by additional ablation studies and more visual results. 

%% file: sec_supp/impact.tex
\section{Broad impact}
The proposed method for 3D generation based on a single reference image offers significant advantages in various fields, including computer graphics, virtual reality, and computer-aided design. One of the main advantages of our method is its ability to produce accurate and detailed 3D models with minimal input data, thereby reducing the need for complex and time-consuming data acquisition processes. This can lead to significant cost savings and increased efficiency in industries such as architecture, entertainment, and manufacturing. Additionally, our approach is likely to be more accessible to non-experts, fostering creativity and innovation in 3D content creation.

However, there are potential limitations to the proposed method. The reliance on a single reference image may result in incomplete or ambiguous 3D reconstructions, particularly in cases where the input image lacks sufficient detail or contains occlusions. In terms of ethical and moral considerations, the adoption of our 3D generation method could raise concerns about privacy and intellectual property rights. We are acutely aware of the potential for our approach to be misused. Therefore, we plan to investigate the implementation of robust watermarks for the generated 3D contents.


%% file: sec_supp/implementation.tex
\section{More implementation details of image-to-3D generation}
\label{sec:implementation}

\subsection{Coarse stage training} 
In the coarse stage, we adopt Instant NGP~\cite{muller2022instant} as the 3D representation. The chosen architecture has a 16-level hash encoding of size $2^{19}$ and entry dim 2. We train the coarse stage from $64$ to $128$ rendering resolution. 

During training, we optimize the Instance NGP with reference view reconstruction loss and a hybrid SDS loss provided by DeepFloyd~\cite{DF} (2D image diffusion model) and Zero-1-to-3~\cite{liu2023zero} (3D novel view synthesis diffusion model). The CFG scale of 2D SDS is set to 20, and we sample time steps from $t\sim \gU(0.2, 0.6)$; For 3D SDS, we set the CFG scale to 5.0 and sample time steps from $t\sim \gU(0.2, 0.5)$. To further regularize the object geometry, we also incorporate a reference view depth loss ~\cite{tang2023make} and normal loss. We train the coarse stage for 3000 iterations, which takes approximately 30 minutes on a 40G A100 GPU. 
\input{tables_supp/algo}
\subsection{Refine stage training} 
During the refine stage, we choose DMTet\cite{shen2021dmtet} as the 3D representation. DMTet is a hybrid SDF-Mesh 3D representation comprising deformable tetrahedral grid $(V_T, T)$ which is capable of differentiable rendering and explicit high-resolution shape modeling. The deformation vector is initialized to $0$ and SDF is initialized by converting the coarse stage density field.  For the texture field, we employ the same setting as the aforementioned Instant NGP. The novel view results can be rendered by a differentiable rasterizer which rasterizes extracted mesh from DMTet and the texture field that gets a 3D intersection from the rasterizer as input. We train the refine stage at the image resolution of $1024$.

During training, we fix the tetrahedral grid and focus on optimizing texture details. We use reference view reconstruction loss and our proposed RGSD loss to optimize the texture field. A summarized algorithm is provided in Algorithm.~\ref{alg:rgsd}. The RGSD loss is provided by a depth-conditioned SD model~\cite{SDD} with CFG scale set to 7.5. We use $T=20$ steps' DDIM inversion, the attention injection start step is set to $l=12$, and we sample time steps $\tau \in [0,l)$ to constrain the difference between intermediate noisy states $\bm{x}_\tau$ and enhanced states $\Tilde{\bm{x}}_\tau$. 
To accelerate training, we pre-select two fixed camera views and derive their $\Tilde{\bm{x}}_0$ states, using them as the optimization target at the $\tau=0$ time step. Then, we alternate between sampling $\tau=0$ to optimize the pre-defined $\Tilde{\bm{x}}_0$ states with fixed camera poses and sampling $\tau \in (0,l)$ to optimize the intermediate $\Tilde{\bm{x}}_\tau$ states with random camera poses. 
In our experiments, we optimize for totally 1000 training iterations in the refine stage, 
which takes about 10 minutes on a 40G A100 GPU.

\subsection{Camera Settings} 
During training, we sample the reference view and random camera views. For the random view sampling, the elevation angles is uniformly sampled from [-10, 60], and the azimuth angle is uniformly sampled from [-180, 180]. We set the FOV fixed to 20 and the camera distance fixed to 3.8 during training.
\input{tables_supp/runtime}
\subsection{Training speed}
We provide a training speed comparison in Tab.~\ref{time}. ``Ours$^{*}$'' represents the settings adopted in the ablation experiment in Section.4.4 of the main text, where we use SDS loss in the refine stage instead of RGSD loss. Compared to the baselines, our method achieves the best results with the least optimization time.

%% file: tables_supp/algo.tex
\begin{algorithm}[tb]
\caption{RGSD loss}
\label{alg:rgsd}
\textbf{Input: }{Depth-conditioned SD model~\cite{SDD} $\epsilon_\phi$, reference image, 3D model with parameter $\theta$, attention injection start step $t=l$, learning rate $\eta$. }

\begin{algorithmic}[1]
\WHILE{not converged}
\STATE Sample camera pose $c$ and render $\bm{x}_0 = g(\theta,c)$
\STATE Sample stop time step $t = \tau$
\STATE \textcolor{blue}{\textbf{\#\textit{DDIM Inversion}}}
\FOR{$t=1,2,...,l$} 
\STATE $\bm{x}_{t} = (\alpha_{t}/\alpha_{t-1})(\bm{x}_{t-1}-\sigma_{t-1}\epsilon_\phi)+\sigma_{t}\epsilon_\phi$ 
\ENDFOR
\STATE \textcolor{blue}{\textbf{\#\textit{DDIM Sampling with Attention Injection}}}
\FOR{$t=l,l-1,...,\tau+1$} 
\STATE $\Tilde{\bm{x}}_{t-1} = (\alpha_{t-1}/\alpha_t)(\Tilde{\bm{x}}_t-\sigma_t\Tilde{\epsilon}_t)+\sigma_{t-1}\Tilde{\epsilon}_t$
\ENDFOR
\STATE Get enhanced state $\Tilde{\bm{x}}_\tau$, noise $\Tilde{\epsilon}_\tau$
\STATE Construct $\bm{x}_\tau = \alpha_\tau\bm{x}_0 + \sigma_\tau\Tilde{\epsilon}_\tau$
\STATE $\theta \leftarrow \theta - \eta\nabla_{\theta} \mathbb{E}[\|\bm {x}_\tau - \Tilde{\bm{x}}_\tau\|_2^2]$
\ENDWHILE
\STATE {\bfseries return} 
\end{algorithmic}
\end{algorithm}

%% file: tables_supp/runtime.tex
\begin{table}[t]
\caption{Training speed comparison.}
\label{time}
	\centering
         \resizebox{1.\textwidth}{!}{
	\begin{tabular}{lccccc}
		\toprule
            Method && Ours & Ours$^{*}$ & Make-It-3D~\cite{tang2023make} & Magic123~\cite{qian2023magic123}\\
        \midrule
        \multirow{2}{*}{Coarse stage}&loss&SDS&SDS&SDS&SDS\\   
        &time(minutes)&30&30&60&30\\
        \midrule
        \multirow{2}{*}{Refine stage}&loss&RGSD&SDS&SDS&SDS(+textual inversion)\\
        &time(minutes)&\textbf{10}&60&60&120+30\\
        \bottomrule
	\end{tabular}
    }
\end{table}

%% file: sec_supp/ablation.tex
\input{images_tex_supp/inversion_example}
\section{Additional ablation studies}
\label{sec:ablation}
\subsection{Effectiveness of depth-based DDIM inversion}
Our proposed RGNV pipeline and RGSD loss are built on the discovery that performing DDIM inversion on a reference image using a depth-conditioned SD model ~\cite{SDD} can significantly improve the reconstruction quality, which enables capturing fine texture details of the reference image in an optimization-free manner. 
As shown in Fig.~\ref{fig:ddim}, compare with regular DDIM inversion, depth-based DDIM inversion can significantly improve the reconstruction quality of the reference images. Compare with the optimization-based Null-text inversion~\cite{mokady2023null}, depth-based DDIM inversion achieves comparable reconstruction quality.
\input{tables_supp/reconstruction}
\input{tables_supp/views}
\input{images_tex_supp/midas}
We further conduct quantitative image reconstruction comparison between regular DDIM inversion and depth-based DDIM inversion using 400 images (introduced in Section.4.2 in the main text), and compute L2 distance and LPIPS~\cite{zhang2018unreasonable}
between the reference image and the reconstructed image. As shown in Tab.~\ref{tab:depth}, the quantitative results demonstrate depth-based DDIM iversion significantly improves the reconstruction quality.
This enables us to obtain an accurate representation of the input image (both high level structure and low level textures) and adapt it to high-fidelity novel-view synthesis in a zero-shot manner.
\input{images_tex_supp/timestep}
\subsection{Robustness for depth condition}
Our RGNV pipeline and RGSD loss relies on a depth-conditioned SD model (SD-depth)~\cite{SDD}, which is originally designed to accept a normalized depth map as input and generates a corresponding color image. 
In our implementation, we mask the estimated depth map using foreground mask of the object, and utilize SD-depth with the masked depth map to provide \textit{shape constraints} for better DDIM inversion and texture transfer, which do not require \textit{accurate depth values}. As demonstrated in Fig.~\ref{midas}, applying the RGNV pipeline to a coarse novel view using estimated depth maps yields an ``Enhanced result A''; In comparison, we manually corrupt the estimated depth maps by averaging its depth value, and make it only provide shape constraints for the RGNV pipelne, the produced ``Enhanced result B'' posses similarly high-quality. It demonstrates the RGNV pipeline and RGSD loss do not rely on an accurate depth estimation module, and are robust for depth conditions. 
\subsection{Robustness for non-frontal views}
We conduct ablations to evaluate the robustness of our method for non-frontal views. Following the experimental settings in Tab.1 (main text) on the GSO~\cite{downs2022google} 3D dataset, we further report the variation of the LPIPS metric in relation to different novel views, ranging from $0^\circ$ (reference view) to $180^\circ$ (back view).
The results are shown in Tab.~\ref{tab:views}. It can be found that the baselines suffer from performance declines when generating novel views deviating from the reference view, but our method still brings performance improvements for each view. It demonstrates the robustness of our method in improving generation quality of both frontal views and non-frontal views.
\subsection{Ablation on attention injection start step}
In the RGNV pipeline,
we perform $t=l$ steps' DDIM inversion to invert the reference image and coarse novel view into noisy states, then perform DDIM sampling with attention injection to transfer fine textures from reference image to the coarse novel view. The impact of different attention injection start step $l$ is shown in Fig.~\ref{fig:timestep}.
We experiment with the commonly used 50 steps' DDIM sampling and inversion in the experiment. It can be observed that as $l$ increases, the texture of the enhanced image approaches that of the reference image more closely, but may introduce geometry change. Therefore, for zero-shot novel view synthesis tasks, we adopt $l=30$ steps. For the implementation of RGSD loss, we use 20 steps' DDIM sampling and inversion, and use $l=12$ for attention injection.
\input{images_tex_supp/comp_syncdreamer}
\input{images_tex_supp/zero_gso}
\input{images_tex_supp/i23_gso}

%% file: images_tex_supp/inversion_example.tex
\begin{figure*}[h]
  \centering
  \includegraphics[width=1.\textwidth]{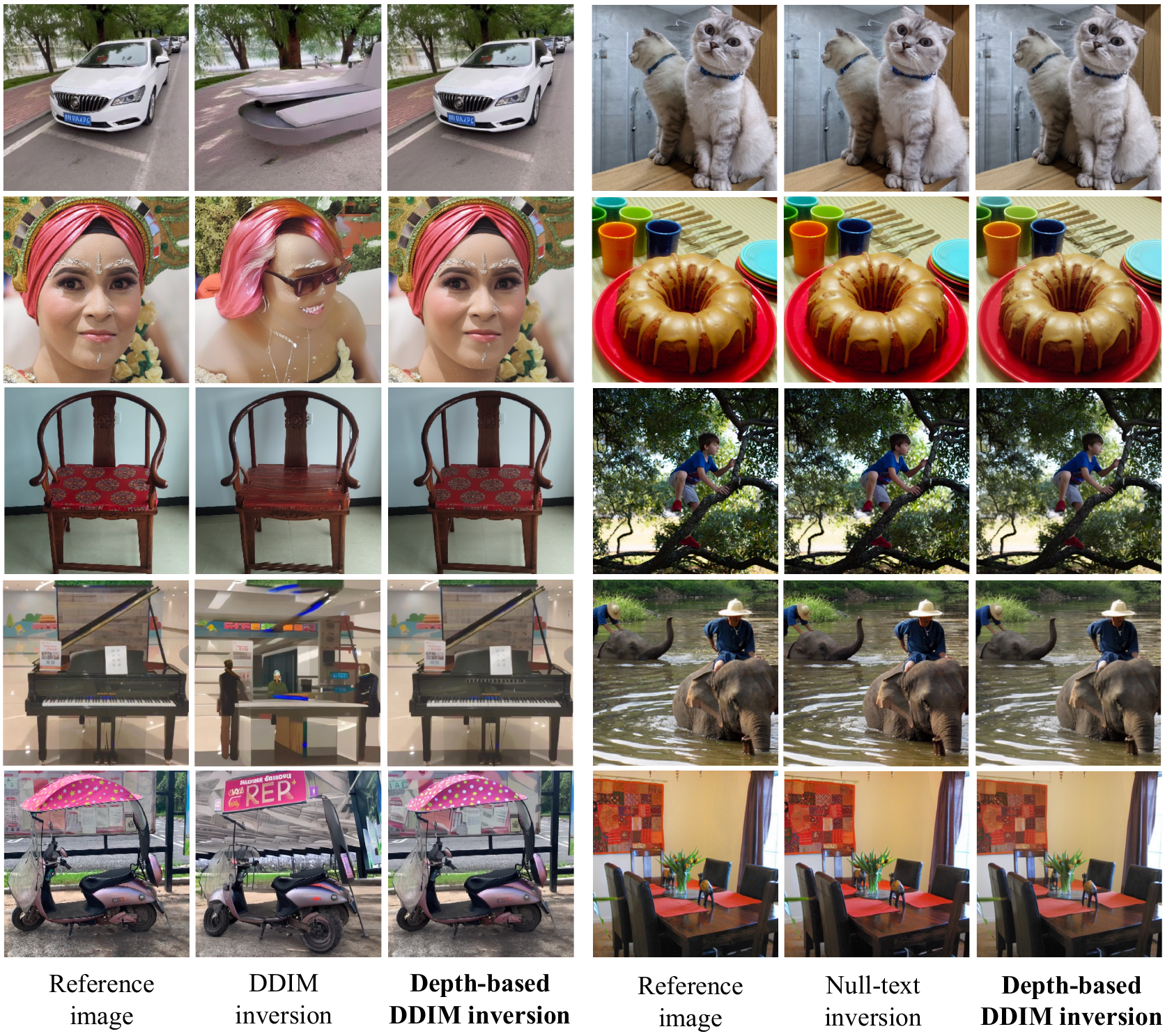}
  \caption{Comparison between depth-based DDIM inversion, regular DDIM inversion and optimization-based Null-text inversion~\cite{mokady2023null}. Example images partly from~\cite{mokady2023null}.} 
\label{fig:ddim}
\end{figure*}

%% file: tables_supp/reconstruction.tex
\begin{table}[t]
  \caption{Quantitative comparison on the reconstruction quality between regular DDIM inversion and depth-based DDIM inversion.}
  \label{tab:depth}
\centering
  \resizebox{0.55\linewidth}{!}{
\begin{tabular}{l|cc}
\toprule
 & LPIPS$\downarrow$  & L2$\downarrow$ \\ \midrule
DDIM inversion& 0.2835 & 159.14 \\
Depth-based DDIM inversion& \textbf{0.0661} & \textbf{66.93}  \\
\bottomrule
\end{tabular}}
\end{table}

%% file: tables_supp/views.tex
\begin{table}[t]
  \caption{Variation of the LPIPS metric in relation to different novel views, ranging from $0^\circ$ (reference view) to $180^\circ$ (back view). Evaluated on the GSO~\cite{downs2022google} 3D dataset adopted in Tab.1 of the main paper.}
  \label{tab:views}
\centering
    \resizebox{.95\linewidth}{!}{
  \begin{tabular}{l|cccccccccccccc}
\toprule
LPIPS$\downarrow$& $0^\circ$ & $22.5^\circ$ & $45^\circ$ & $67.5^\circ$ & $90^\circ$ & $112.5^\circ$ & $135^\circ$ & $157.5^\circ$ & $180^\circ$ \\ \midrule
Zero-1-to-3& 0.051&0.120&0.146&0.178&0.179&0.183&0.188&0.189&0.188\\
\textbf{Zero-1-to-3+Ours}& 0.047&0.112&0.132&0.158&0.162&0.167&0.170&0.172&0.172\\
Syncdreamer& 0.065&0.103&0.129&0.145&0.153&0.159&0.166&0.168&0.168\\
\textbf{Syncdreamer+Ours}& 0.061&0.097&0.111&0.122&0.125&0.131&0.138&0.140&0.140\\
\bottomrule
\end{tabular}}
\end{table}

%% file: images_tex_supp/midas.tex
\begin{figure}[t]
  \centering
  \includegraphics[width=1.\textwidth]{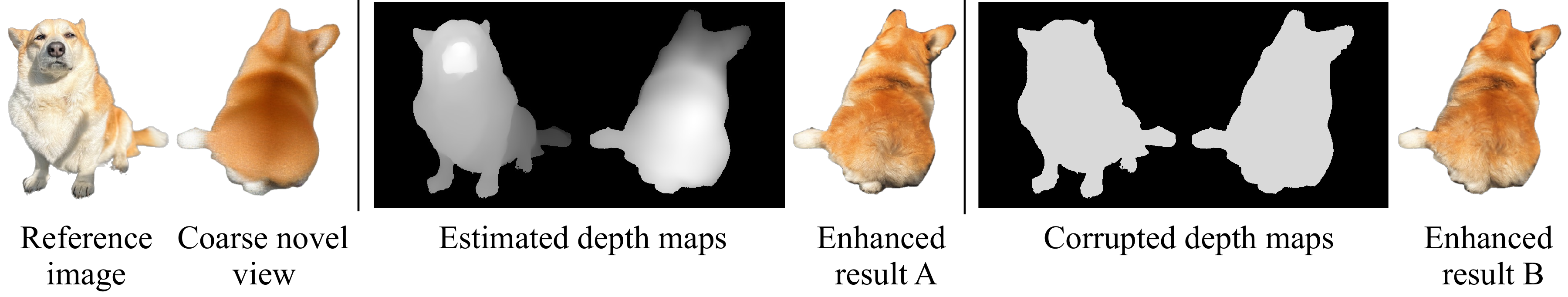}
  \caption{Robustness for depth condition.}
  \label{midas}
\end{figure}

%% file: images_tex_supp/timestep.tex
\begin{figure*}[t]
  \centering
  \includegraphics[width=1.\textwidth]{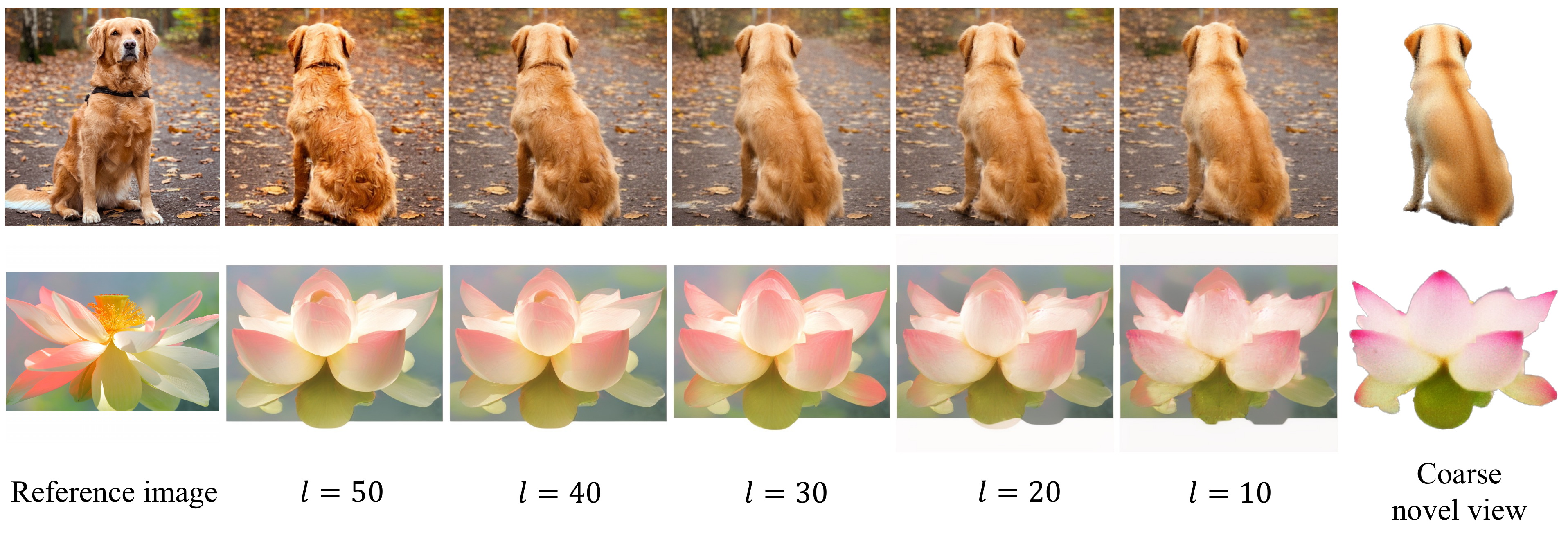}
  \caption{Ablation on attention injection start time step $l$.}
\label{fig:timestep}
\end{figure*}

%% file: images_tex_supp/comp_syncdreamer.tex
\begin{figure*}
  \centering
  \includegraphics[width=1.\textwidth]{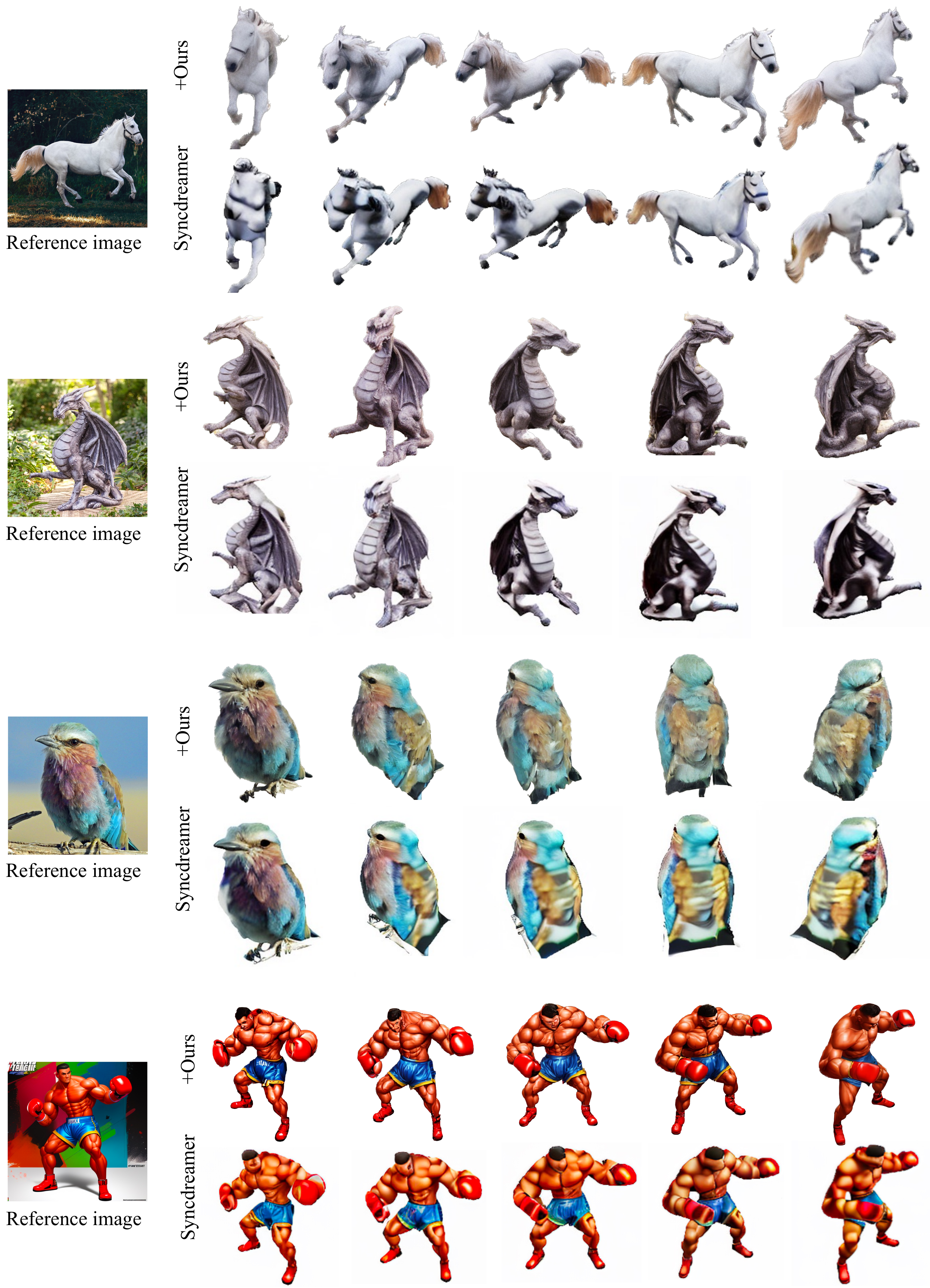}
  \caption{Qualitative comparison with Syncdreamer. It can be found that our method can generate novel views with higher fidelity according to the reference image.}
\label{fig:comp_sync}
\end{figure*}

%% file: images_tex_supp/zero_gso.tex
\begin{figure*}[htb]
  \centering
  \includegraphics[width=1.\textwidth]{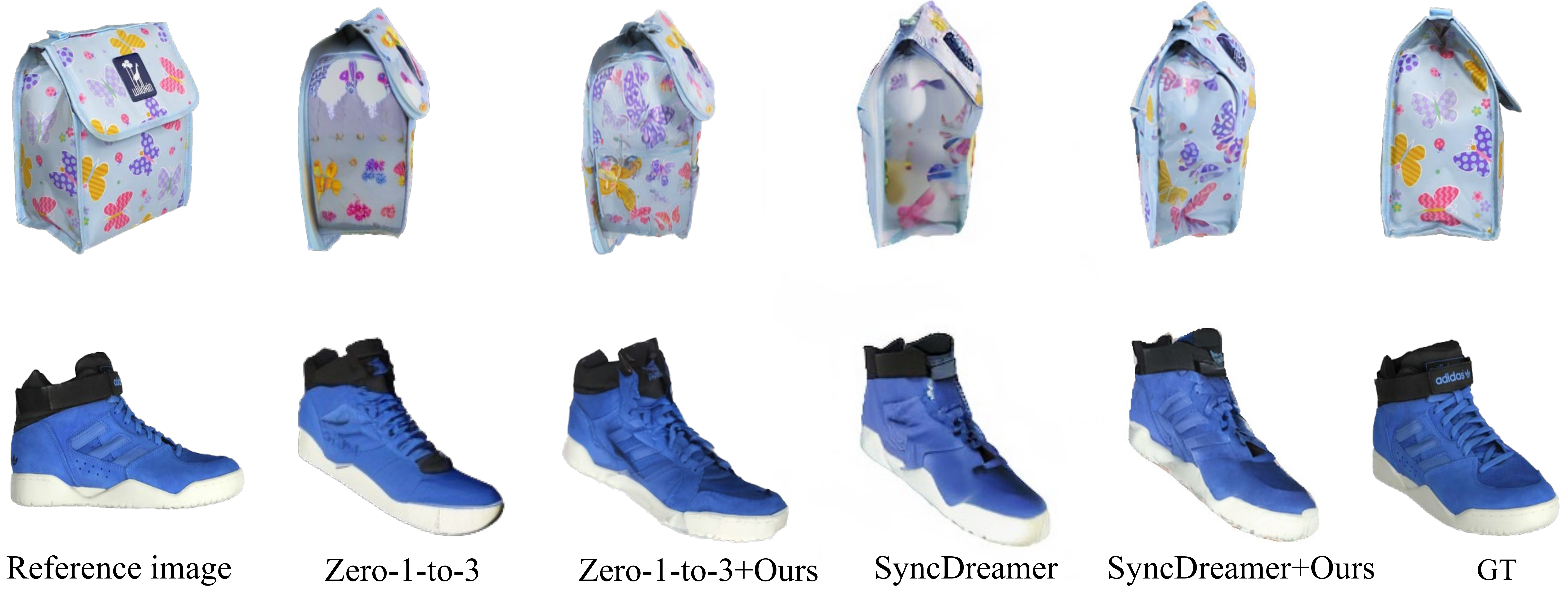}
  \caption{Qualitative comparison with Zero-1-to-3~\cite{liu2023zero} and SyncDreamer~\cite{liu2023syncdreamer} on GSO dataset.}
\label{fig:zero_gso}
\end{figure*}

%% file: images_tex_supp/i23_gso.tex
\begin{figure*}[htb]
  \centering
  \includegraphics[width=.81\textwidth]{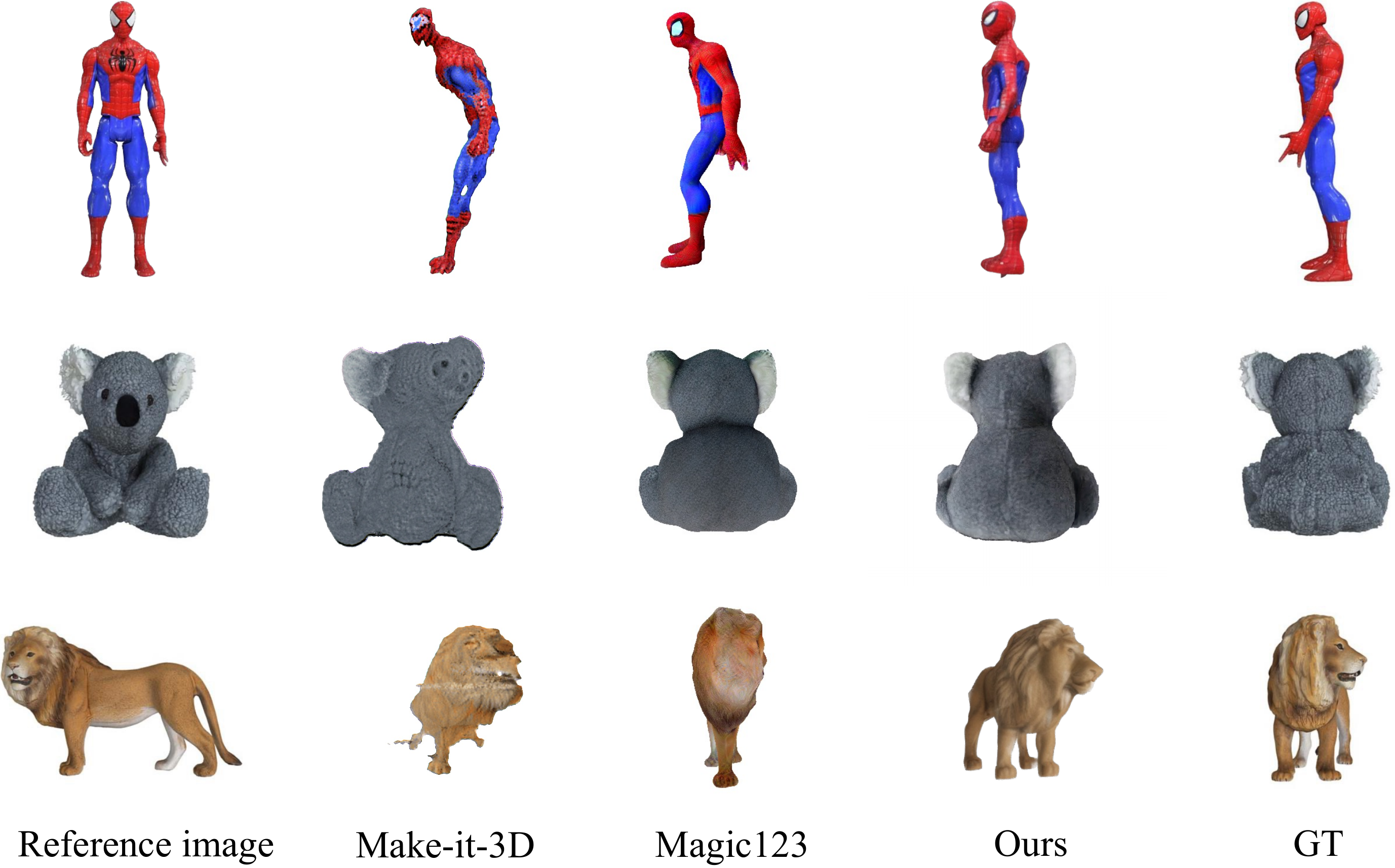}
  \caption{Qualitative comparison with image-to-3D generation baselines on GSO dataset.}
\label{fig:i23_gso}
\end{figure*}

%% file: sec_supp/moreresults.tex
\section{More qualitative results}
\subsection{Qualitative comparisons in the main text}
The qualitative comparisons with SyncDreamer~\cite{liu2023syncdreamer} (introduced in Section.4.2 in the main text) on zero-shot novel view synthesis are shown in Fig.~\ref{fig:comp_sync}. 
We removed the synthesized background of our method for a more direct comparison. The results show that our method can generated novel views more consistent with the
reference image, and demonstrate that it can be seamlessly used to improve visual quality of different zero-shot novel view synthesis methods~\cite{liu2023syncdreamer,liu2023zero}.

The qualitative comparisons on GSO dataset~\cite{downs2022google} (introduced in Section.4.2 and Section.4.3 in the main text) are shown in Fig.~\ref{fig:zero_gso} and Fig.~\ref{fig:i23_gso}. In zero-shot novel view synthesis (Fig.~\ref{fig:zero_gso}), our method produces novel views with more consistent texture according to the reference image. In image-to-3D generation (Fig.~\ref{fig:i23_gso}), our method can generate reasonable geometry and consistent textures compared with baselines. Quantitative comparisons are reported in Tab.1 and Tab.2 of the main text.

\subsection{More synthetic results}
We present more image-to-3D generation results of out method, video results are available at the Supplementary project page.
\input{./images_tex_supp/teaser2}
\input{./images_tex_supp/teaser3}
\input{./images_tex_supp/teaser4}

%% file: images_tex_supp/teaser2.tex
\begin{figure*}
\centering
    \includegraphics[width=\linewidth]{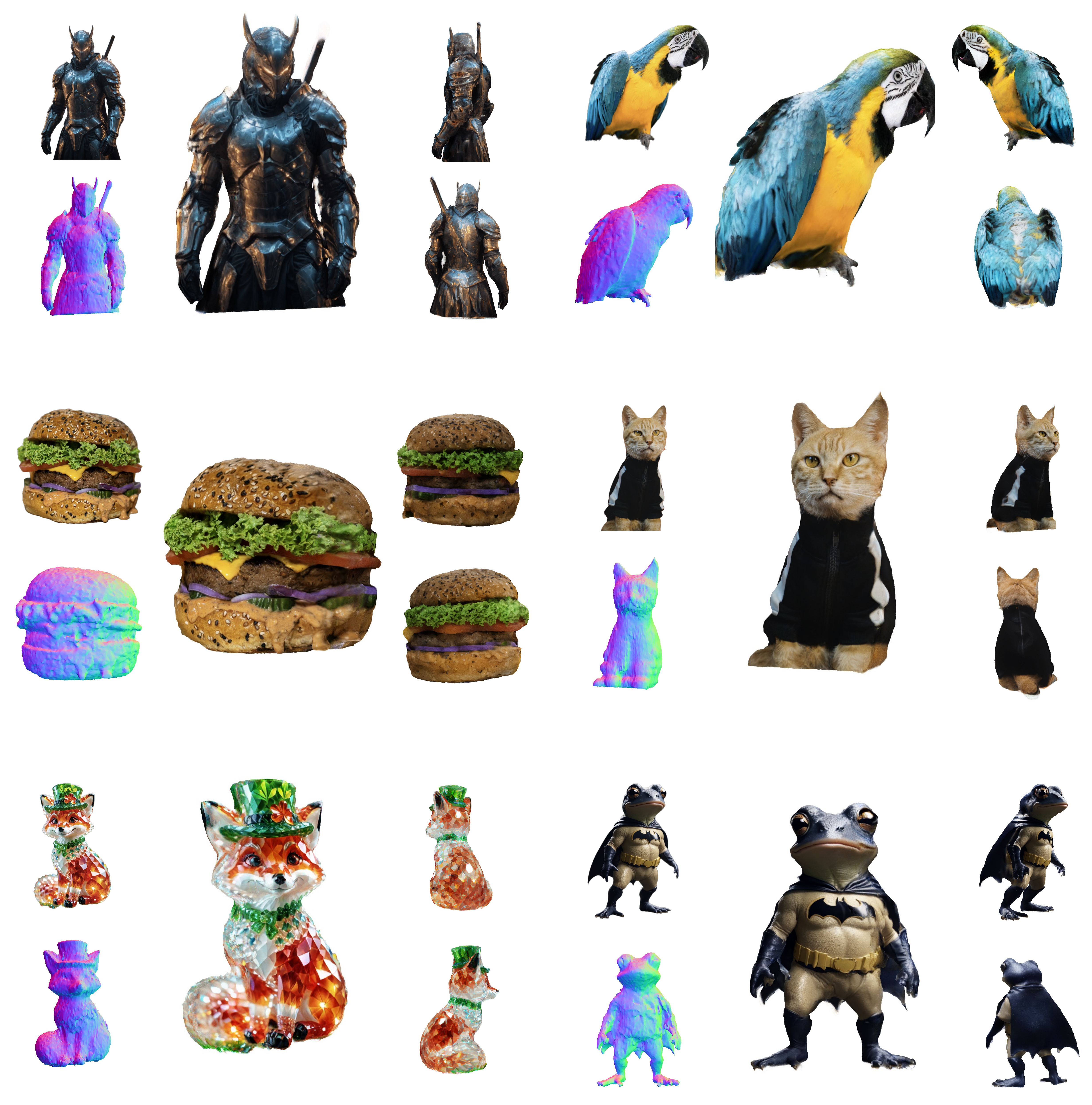}
    \vspace{.8cm}
    \caption{More image-to-3D generation results. In each block above, we display the reference image (top left corner) along with the rendered novel views and normal of the generated 3D content. The presented novel views demonstrate that our approach maintains consistency and high-fidelity  with the reference image, even in views significantly deviating from the reference view.}
\end{figure*}

%% file: images_tex_supp/teaser3.tex
\begin{figure*}
\centering
    \includegraphics[width=\linewidth]{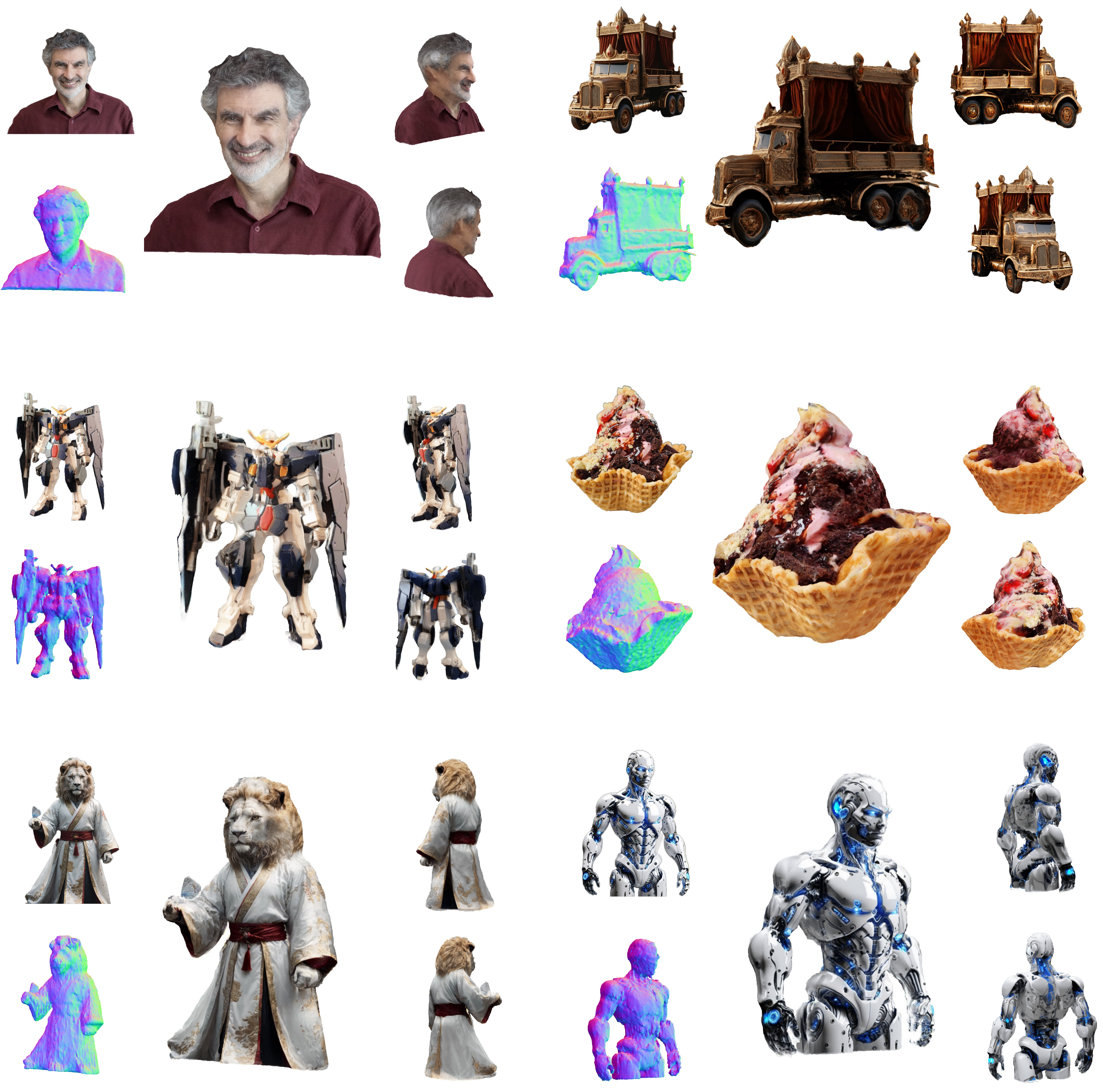}
    \vspace{.8cm}
    \caption{More image-to-3D generation results. In each block above, we display the reference image (top left corner) along with the rendered novel views and normal of the generated 3D content. The presented novel views demonstrate that our approach maintains consistency and high-fidelity  with the reference image, even in views significantly deviating from the reference view.}
\end{figure*}

%% file: images_tex_supp/teaser4.tex
\begin{figure*}
\centering
    \includegraphics[width=\linewidth]{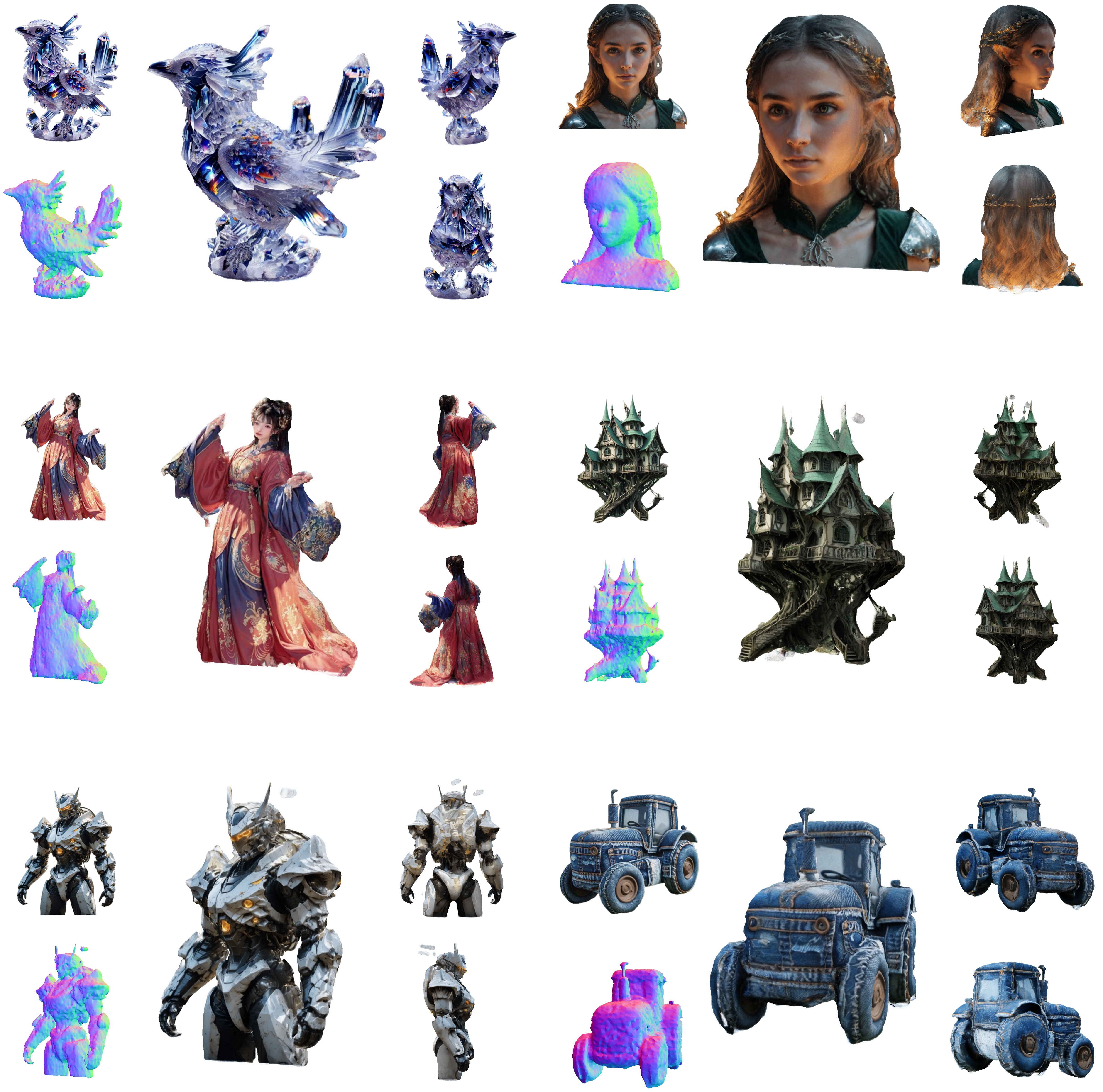}
    \vspace{.8cm}
    \caption{More image-to-3D generation results. In each block above, we display the reference image (top left corner) along with the rendered novel views and normal of the generated 3D content. The presented novel views demonstrate that our approach maintains consistency and high-fidelity  with the reference image, even in views significantly deviating from the reference view.}
\end{figure*}